\documentclass[12pt]{elsarticle}

\usepackage{amssymb}
\usepackage{amsmath}
\usepackage{xcolor}
\usepackage{colortbl}
\definecolor{control}{RGB}{173,216,230} 
\definecolor{lightgray}{gray}{0.9}
\definecolor{lightblue}{rgb}{0.8,0.9,1}
\usepackage{caption}
\usepackage{subcaption}
\usepackage{hyperref}
\usepackage{amsfonts}
\usepackage[utf8]{inputenc}
\usepackage[T1]{fontenc}
\usepackage{chemformula} 
\usepackage{pdfpages}

\begin{document}

\begin{frontmatter}

\journal{arXiv}

\title{PROTECT: Protein circadian time prediction using unsupervised learning}

\author{Aram Ansary Ogholbake}
\author{Qiang Cheng \corref{cor1}}
\cortext[cor1]{Corresponding author. Email: qiang.cheng@uky.edu}
\address{Institute for Biomedical Informatics, Department of Internal Medicine and Department of Computer Science, University of Kentucky, Lexington, KY, {US}}

\begin{abstract}
Circadian rhythms regulate the physiology and behavior of humans and animals. Despite advancements in understanding these rhythms and predicting circadian phases at the transcriptional level, predicting circadian phases from proteomic data remains elusive. This challenge is largely due to the scarcity of time labels in proteomic datasets, which are often characterized by small sample sizes, high dimensionality, and significant noise. Furthermore, existing methods for predicting circadian phases from transcriptomic data typically rely on prior knowledge of known rhythmic genes, making them unsuitable for proteomic datasets.
To address this gap, we developed a novel computational method using unsupervised deep learning techniques to predict circadian sample phases from proteomic data without requiring time labels or prior knowledge of proteins or genes. Our model involves a two-stage training process optimized for robust circadian phase prediction: an initial greedy one-layer-at-a-time pre-training which generates informative initial parameters followed by fine-tuning. During fine-tuning, a specialized loss function guides the model to align protein expression levels with circadian patterns, enabling it to accurately capture the underlying rhythmic structure within the data. We tested our method on both time-labeled and unlabeled proteomic data. For labeled data, we compared our predictions to the known time labels, achieving high accuracy, while for unlabeled human datasets, including postmortem brain regions and urine samples, we explored circadian disruptions. Notably, our analysis identified disruptions in rhythmic proteins between Alzheimer's disease and control subjects across these samples.
\end{abstract}

\begin{keyword}
Circadian Rhythms \sep Proteomic Data \sep Alzheimer's Disease

\end{keyword}
\end{frontmatter}

\color{black}

\section{Introduction}

Circadian rhythms, driven by an internal molecular clock, regulate physiological and behavioral processes. Disruptions to these rhythms have been associated with various pathologies, including type 2 diabetes, obesity, and neurological disorders like Alzheimer's disease (AD) \cite{leng2019association,logan2019rhythms,stenvers2019circadian,teeple2023high,javeed2018circadian,jarrett1969diurnal}. Hence, studying these rhythms is crucial due to their significant impact on health and disease.

In mammals, cell-autonomous circadian rhythms are driven by interconnected transcriptional-translational feedback loops. Additionally, substantial contributions to circadian processes come from posttranslational and posttranscriptional regulation \cite{mauvoisin2020proteomics,feng2012clocks}. While the majority of research in the field of chronobiology has focused on the rhythmic expression at the mRNA level, understanding circadian rhythmicity at the protein level remains limited. This limitation is noteworthy given the acknowledged contribution of post-transcriptional mechanisms to circadian rhythms at the protein level \cite{kojima2011post,robles2014vivo}.

Obtaining precise sample times is currently essential for analyzing circadian rhythms in proteomic data. 
However, many datasets, particularly human samples, often lack labeled timestamps due to constraints imposed by health risks in collection protocols or the inability to collect precise sample times (e.g., time of death).  This lack of time labels poses an urgent need for computational methods to predict the phase of each sample. Challenges arise in developing such methods due to several factors. Firstly, these datasets often contain small sample sizes with high dimensionality and significant noise. Moreover, the presence of proteins with periods shorter than 24 hours, i.e., ultradian rhythms, adds complexity. These challenges cannot be effectively tackled using conventional statistical or machine learning methods. The rapid advances in deep learning techniques offer promise for overcoming these challenges and estimating the circadian phase of each sample. While efforts have been made at the mRNA level \cite{anafi2017cyclops,braun2018universal,leng2015oscope,hughey2016zeitzeiger,vlachou2024timeteller,duan2024taufisher,woelders2023machine}, there is a notable gap in research concerning the prediction of circadian phases in proteomic data.

 As mentioned earlier, many datasets lack labeled timestamps. Zeitzeiger \cite{hughey2016zeitzeiger}, TimeSignature \cite{braun2018universal}, TimeTeller \cite{vlachou2024timeteller}, tauFisher \cite{duan2024taufisher}, and PLSR \cite{woelders2023machine}
employ supervised learning approaches that require time labels and, therefore, cannot be used in practice when time information is unavailable. In contrast, we introduce a novel unsupervised learning approach which eliminates the need for time labeled samples to estimate circadian phases. This makes it a valuable alternative for circadian rhythm analysis, particularly in scenarios where accurate time annotations are lacking. \color{black}

On the other hand, existing unsupervised or mathematical sample phase estimation (or temporal ordering) methods for gene expression data, such as CYCLOPS \cite{anafi2017cyclops} and CIRCUST \cite{larriba2023circust}, were specifically developed for gene expression datasets \footnote{We also tested a program of unpublished ESOCVD \cite{ji4079104esocvd}, which could not execute.} . These methods explicitly require the use of expressions of a set of “seed rhythmic genes” to operate. The seed genes are set of known circadian rhythmic genes, including core clock genes, in animal tissues. Without these pre-selected seed rhythmic genes, these phase estimation methods may not function properly, limiting their applicability to datasets where such prior knowledge is available.

Proteomic data poses a significant challenge for applying these methods to study rhythmic proteins, as many proteins corresponding to the designated seed rhythmic genes may not be expressed in a tissue. This is because proteomic datasets typically measure the expression levels of 1,000 – 10,000 proteins, which is significantly fewer than the 15,000 - 50,000 genes typically measured in gene expression datasets. Furthermore, proteins corresponding to core clock genes are often expressed at low levels or not at all, making their measurements unavailable in proteomic datasets. Additionally, some genes known to be rhythmic at the mRNA level may not exhibit rhythmicity in their corresponding proteins, and vice versa. Previous studies have shown that only a small proportion of rhythmic proteins are rhythmic in their corresponding genes, while a large number of proteins exhibit rhythmicity even when their corresponding mRNAs do not, and vice versa \cite{mauvoisin2014circadian}.

To meet the urgent need and challenges, we introduce an unsupervised deep learning approach called PROTECT (PROTEin Circadian Time prediction). PROTECT is a rhythmicity-aware model designed to predict the circadian phase of each sample in proteomic data, without requiring time labels or prior knowledge of rhythmic markers. This method can effectively handle small sample size datasets and ultradian proteins. It does not require any known seed rhythmic proteins or genes and can handle considerable noise levels present in proteomic data. We demonstrate the efficacy, accuracy, and robustness of our approach using mouse, Ostreococcus tauri cell and human datasets, where time labels are available. Subsequently, we investigate circadian rhythms in human datasets obtained from different brain regions and urine samples from both control and AD subjects.

Proteins, along with the metabolic pathways they modulate, often serve as the ultimate biological effectors of AD genetic  \cite{johnson2022large}. Despite the increasing body of work in recent years on discerning disparities in proteomic data between AD and control subjects \cite{seyfried2017multi,andreev2012label,hondius2016profiling,johnson2018deep,haytural2021insights}, there is a lack of investigation in identifying differences between control and AD individuals based on circadian rhythms. Our developed approach can uncover AD-associated rhythmic proteins and distinguish differences in rhythmic patterns between control and AD subjects, filling this gap. 

In brief, the novelty of our paper includes but is not limited to:

\begin{itemize}
    \item We present a unique approach for accurately predicting circadian phases in un-labeled proteomic datasets, which, to our knowledge, no existing methods in the field address. It does not rely on prior information about circadian rhythmic genes or proteins, marking a significant advancement in the field of computational inference from proteomic data.
    \item  Our method shows high prediction accuracy while effectively handling datasets with varying sample sizes, high dimensionality, and substantial noise. Additionally, it identifies ultradian rhythmic proteins, showcasing its versatility.
    \item Applying our method to un-labeled human proteomic datasets reveals circadian rhythmic differences between control and AD subjects across three postmortem brain regions and urine samples. This highlights the potential of our approach to uncover novel insights into circadian disruptions associated with AD in proteomic datasets.
\end{itemize}

\section{Proposed PROTECT model and algorithm}
We developed PROTECT, an unsupervised learning method, to predict the time of high-dimensional proteomic samples based on the data itself, without relying on any a priori information or time labels. Proteomic data includes both rhythmic and non-rhythmic proteins. Rhythmic proteins typically exhibit periodic patterns, with values peaking at certain times of the day and dipping at others, while non-rhythmic proteins have no periodic patterns. Moreover, the peak times among rhythmic proteins are different. Without known sample times, rhythmic patterns are obscured, and PROTECT aims to recover them. \color{black} PROTECT addresses key challenges in studying circadian rhythms using proteomic data,  including small sample sizes, the presence of ultradian proteins, and limited knowledge of rhythmic proteins, especially in human datasets. 

PROTECT utilizes a deep neural network (DNN) to predict the phase of each sample in proteomic data using a greedy layer-wise technique, inspired by Hinton et al.'s work \cite{hinton2006fast}. The DNN architecture in PROTECT is well-suited for high-dimensional data. It consists of an input layer, multiple hidden layers, and an output layer with two neurons representing a single angular phase. Training the network involves pre-training the DNN and obtaining the weights of the corresponding DNN neurons, followed by fine-tuning the weights to predict sample phases by regressing with proteins' cosine models. 
\begin{figure}
\centering
\includegraphics[scale=0.103]{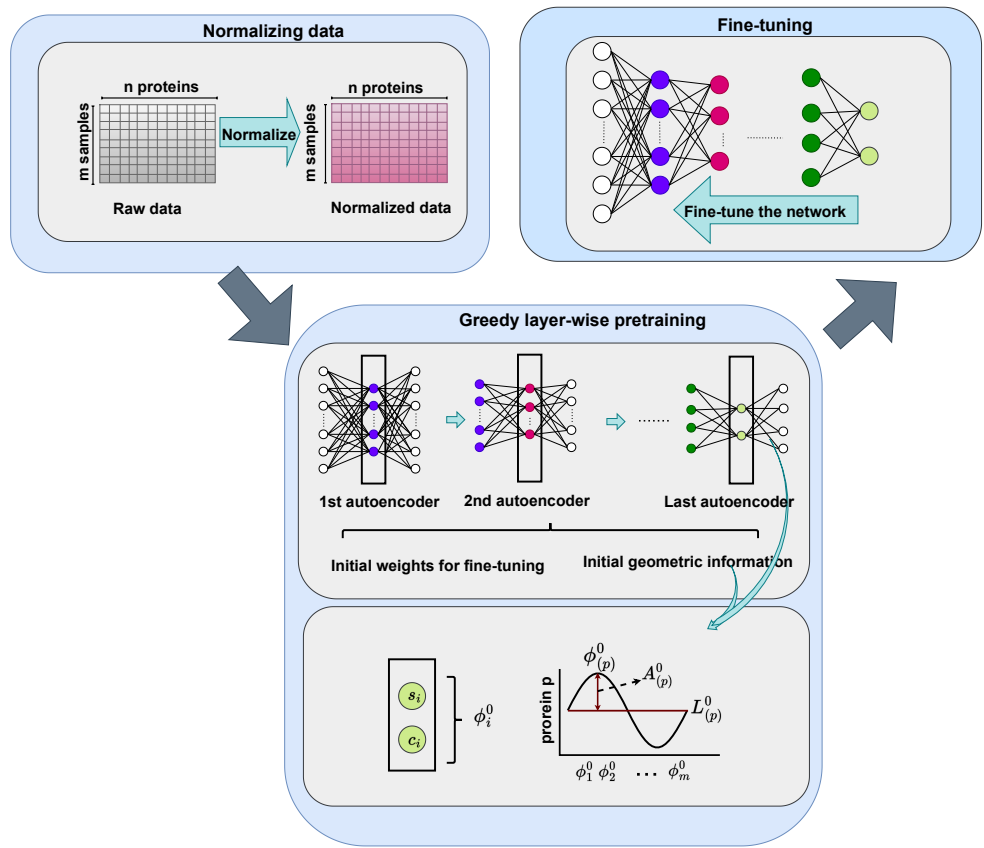}
\caption{Illustration of the overall diagram of PROTECT.}
\label{fig:illustration}
\end{figure}

Greedy layer-wise reconstruction has advantages shown in \cite{hinton2006fast}, which we adapt for pre-training our DNN architecture. The optimized weights obtained in the pre-training stage provide an effective initialization, facilitating effective optimization for learning the sample phases and fitting downstream cosine models in the fine-tuning stage. This approach is particularly effective for small datasets because of its robust initialization of the network's weights, which can be crucial when data is limited. Additionally, by training each layer independently, the network captures features at different levels of abstraction, reducing the risk of overfitting that often accompanies small sample sizes. The hierarchical representation further augments the network's capability to capture intricate patterns within the high-dimensional proteomic data.

PROTECT's methodology is structured into three main steps, as depicted in Figure \ref{fig:illustration}: Normalizing data, Greedy layer-wise pre-training, and Fine-tuning.

\textbf{Normalizing data: }In the first step, the proteomic data, which includes $m$ samples with $n$ protein measurements per sample, undergoes z-score normalization. This involves subtracting the mean and dividing by the standard deviation for each protein across all samples. This standardization ensures that all proteins are on a comparable scale, improving the stability and performance during the training process. The resulting normalized data is then utilized in the subsequent step: Greedy layer-wise pre-training.

For datasets with a very large number of features (e.g.,$n\geq5000$), we recommend applying a feature selection or dimensionality reduction technique after normalizing the data to enhance computational efficiency and model performance. One effective approach is to use k-means clustering, where the features are grouped into clusters, and only the clusters with the highest variance are kept. Alternatively, high-variance features can be directly selected, where only the features with the highest variance across samples are chosen.\color{black}

\textbf{Greedy layer-wise pre-training: }This step involves greedily training each layer of the DNN using a separate shallow auto-encoder (AE) network suitable for datasets with small sample sizes. The primary objective is to encode the input into lower dimensions and capture intricate patterns in the proteomic data using each AE, ultimately reaching the output layer where the features represent the encoded data in two dimensions (see an example of encoded data in supplementary Figure S1). \color{black}

Each shallow auto-encoder consists of an input layer, a hidden layer, and an output layer. The hidden layer's output from the AE at layer $l$ of the DNN becomes the input for the AE at layer $l+1$ in the DNN. The objective function for each AE is the mean squared error (MSE) between the input and output, providing a measure of how well it reconstructs the input. 

After training the last AE, the values of the two nodes in its hidden layer, denoted as $s_i$ and $c_i$ for sample $i$, are extracted. These feature values compute the initial phase, $\phi^{0}_i$, for each sample $i$, as expressed by the equation:
\begin{equation}
\label{eq:phase}
\phi^{0}_i = \arctan \left(\frac{s_i}{c_i}\right).
\end{equation}

Subsequently, CosinorPy is employed to extract geometric information for each protein using the initial predicted phases ($\phi^{0}_i$). This includes parameters such as amplitude ($A^0_{p}$), mesor ($L^0_{p}$), and acrophase ($\phi^0_{p}$), achieved by fitting each protein $p$ to a cosine curve using the initial predicted sample phases. This geometric information is then incorporated into the objective function of the DNN in the next step, which is fine-tuning.

In summary, the pre-training process begins with training the first AE, where normalized proteomic data serves as the input. From this, we extract the hidden layer values. Then, we proceed to train the second AE, utilizing the hidden layer output of the first AE as its input, and extract the hidden values. This iterative process continues until we extract the features of the last AE, which yields the initial angular phase. The initial phase is used to calculate the initial geometrical information of each protein which will be used in fine-tuning. Moreover, the extracted features from each AE are then utilized to initialize the weights of the DNN in fine-tuning.

\textbf{Fine-tuning: }After pre-training the DNN using sequence of shallow AEs and obtaining the initial weights and geometrical information, we fine-tune the network to predict sample phases. In this stage, the network learns to align protein expressions with rhythmicity by fitting each protein to a cosine function. We organize the data into a table where each row corresponds to a sample, and each column represents a proteins's expression; that is, $x_{ip}$ represents the original observation for sample $i$ and protein $p$. We use a parameterized cosine function to fit the observations, given by:
\begin{equation}
    \hat x_{ip} =  L_p + A_p \cos({\omega_{p}\hat \phi_i} + \phi_p),
\end{equation}
where $\hat \phi_i$ represents the sample phase, and $L_p$, $A_p$, and $\phi_p$ are the mesor, amplitude, and the acrophase of protein $p$, respectively. $2\pi /\omega_{p}$ represents the period of protein p.
The objective function is defined as:
\begin{equation}
\label{eq:loss}
\mathfrak{L} = \frac{1}{m}
\sum_{i=1}^{m} \frac{1}{n}\sum_{p=1}^{n}
\| x_{ip} -\hat x_{ip} \|^q_q + \lambda R(\Theta). 
\end{equation}
\noindent Here, $\|\cdot\|_q$ is an $l_q$ norm with respect to both $i$ and $p$, with $q$ as a positive value. $\Theta$ represents the set of all relevant parameters (including  $L_p$, $A_p$, $\hat \phi_i$ and $\phi_p$), and $R(\Theta)$ represents a regularization function on the parameters. $\lambda$ is a non-negative hyperparameter balancing the fitting error and regularization.

The learnable parameters $L_p$, $A_p$, and $\phi_p$ are initialized with the values calculated during the pre-training step.  $\omega_{p}$ allows the model to fit proteins with different rhythmicity such as circadian rhythms and ultradian rhythms.  This flexibility ensures that the model accurately captures various rhythmic patterns in protein expression.

This objective function aims to find the optimal phase for each sample and determines the best geometrical parameters, ensuring an accurate and effective fitting of the protein data to the cosine curve. Moreover, by incorporating the unique geometric information of each protein in each sample, this objective function helps with handling small sample sizes  (to see an example of a random protein recovering its rhythmicity during this stage, see supplementary Figure S2). \color{black}

We have explored the use of multiple regularization functions including $l_1$ norm, $l_2$ norm and total variation (TV) regularization. The total variation term aims to reduce the abrupt shifts between consecutive sample phases and is defined as:
\begin{equation}
     \sum_{k=2}^{m} | \hat \phi_{i_k} - \hat \phi_{i_{(k-1)}} |,
\end{equation}
where sample phases are sorted in ascending order, and $i_k$ represents the k-th sorted sample for $k=1, \cdots, m$.

 Our experiments with various $\lambda$ values for different regularization functions on labeled datasets showed no significant improvement in phase prediction accuracy when $\lambda > 0$. Therefore, we set  the $\lambda = 0$ in our final model. Moreover, we employed $q=1$ for the fitting error term.

This parametric objective function enables the predicted phases to take into account the protein geometrical information, such as amplitudes, acrophases, and periods.\ Moreover, it is applicable when the noise in the data is non-Gaussian \cite{thaben2014detecting,movskon2020cosinorpy}. In addition, cosine curve fitting has been shown to be effective on data without replicates, containing outliers, irregularly spaced time intervals, and unbalanced data distributions where more samples are collected at certain times of the day \cite{movskon2020cosinorpy,ruben2018database,ruben2019large}  To see the performance of the objective function, refer to the convergence results provided in supplementary Figures S3, S4, and S5).\color{black}

\subsection{Screening for potential outliers for samples and proteins}\label{sec: outlier}

In the fine-tuning phase, we design our model to handle inherent noise and potential outliers in proteomic data. The process of detecting outliers is as follows:
\begin{itemize}
    \item \textbf{Sample-level outlier detection:} For each sample, we calculate the averaged fitting error using: 
    \begin{equation}
    E_i = \frac{1}{n}\sum_{p=1}^{n}
 (x_{ip} -\hat x_{ip}) , \quad i=1, \cdots, m.
\end{equation}
Thus, $E_i$ represents the fitting residues averaged for all proteins $p=1, ..., n$.   We then compute the mean and standard deviation of these $E_i$ values across all samples, respectively denoted by $m_s$ and $\sigma_s$. If a sample’s deviation exceeds two standard deviations from the mean $m_s$, i.e., if $|E_i - m_s| > 2 \sigma_s$, then sample $i$ is considered an outlier.
    \item \textbf{Protein-level outlier detection:}  For each protein, we obtain its average fitting residual over all samples: 
     \begin{equation}
         D_p = \frac{1}{m}\sum_{i=1}^{m}
 (x_{ip} -\hat x_{ip}) , \quad p=1, \cdots, n.
\end{equation}
Thus, $D_p$ represents the fitting quality averaged for all samples $i=1, ..., m$. 
We compute the mean $m_p$  and standard deviation $\sigma_p$ of $D_p$ across all proteins. If a protein's deviation exceeds two standard deviations from the mean $m_p$, i.e., if $|D_p - m_p| > 2 \sigma_p$, then protein $p$ is considered an outlier candidate. To ensure no significant cyclic proteins are removed, the outlier candidates undergo a second screening process. In this screening, if the predicted amplitude of a candidate falls below the 75th percentile of all predicted amplitudes, it is classified as an outlier protein.
\end{itemize}
Through this approach, we identify outliers (resp. outlier proteins) that significantly deviate from the majority of samples (resp. proteins). 
Subsequently, we remove the outlier samples and proteins to retrain the model. 
This process minimizes the influence of outlier or corrupted proteins and samples that have unusually large residuals.

\section{Results}
\subsection{Comprehensive Evaluation of PROTECT}
To assess the efficacy of PROTECT, we performed computational experiments on multiple public datasets. We verified the accuracy of our method using labeled proteomic datasets from human, mouse and cell models. Subsequently, we conducted experiments on unlabeled human datasets, including postmortem brains and urine. Our investigation on unlabeled human datasets focused on comparing differences between control and AD subjects. This comprehensive, multifaceted approach enabled us to evaluate PROTECT's capabilities on different species and varying sample sizes datasets.
\subsection{Hyperparameters and Model Details} Our model was trained using PyTorch Lightning, where the best results were obtained using 5 hidden layers. The neuron counts of these layers ranged from $2^{\lfloor \log_2(f) \rfloor}$ to $2$, where $f$ is the number of features in each dataset. During pre-training, optimizers such as SGD, Adam, and learning rate-free methods (DAdapt-SGD and DAdapt-Adam \cite{defazio2023learning}) were tested. We found that the specific optimizer did not notably impact the end results, allowing flexibility in optimizer choice for each layer. During pre-training, the first 5 AEs were trained using 7 epochs and the last AE was trained for 20 epochs. During fine-tuning the whole network was trained for 20 epochs. To handle noise, we retrained the network for 200 epochs using the datasets with outliers removed. Check table \ref{tab:hp} for some details on the implementation.

\begin{table}[h!]
\centering
\small
\begin{tabular}{|p{2.3cm}|p{2.1cm}|p{2.5cm}|p{2cm}|p{2.cm}|}
\hline
\textbf{Training Stage} & \textbf{Optimizer} & \textbf{Best Learning Rate} & \textbf{Momentum} & \textbf{Weight Initialization} \\
\hline
Pre-training (1st-5th AE) & \begin{tabular}[c]{@{}l@{}}Adam \\ SGD \\ DAdapt-SGD\end{tabular} & \begin{tabular}[c]{@{}l@{}}Adam: 0.001 \\ SGD: 0.1 \\ DAdapt-SGD: -\end{tabular} & SGD: 0.85 & Xavier Uniform \\
\hline
Pre-training (last AE) & DAdapt-SGD & learning rate-free & - & Xavier Uniform \\
\hline
Fine-tuning & DAdapt-SGD & learning rate-free & - & Transferred Weights \\
\hline
\end{tabular}
\caption{Details for pre-training and fine-tuning stages.}
\label{tab:hp}
\end{table}

\begin{figure*}
\centering
\begin{tabular}{cccc}
\includegraphics[width=.22\textwidth]{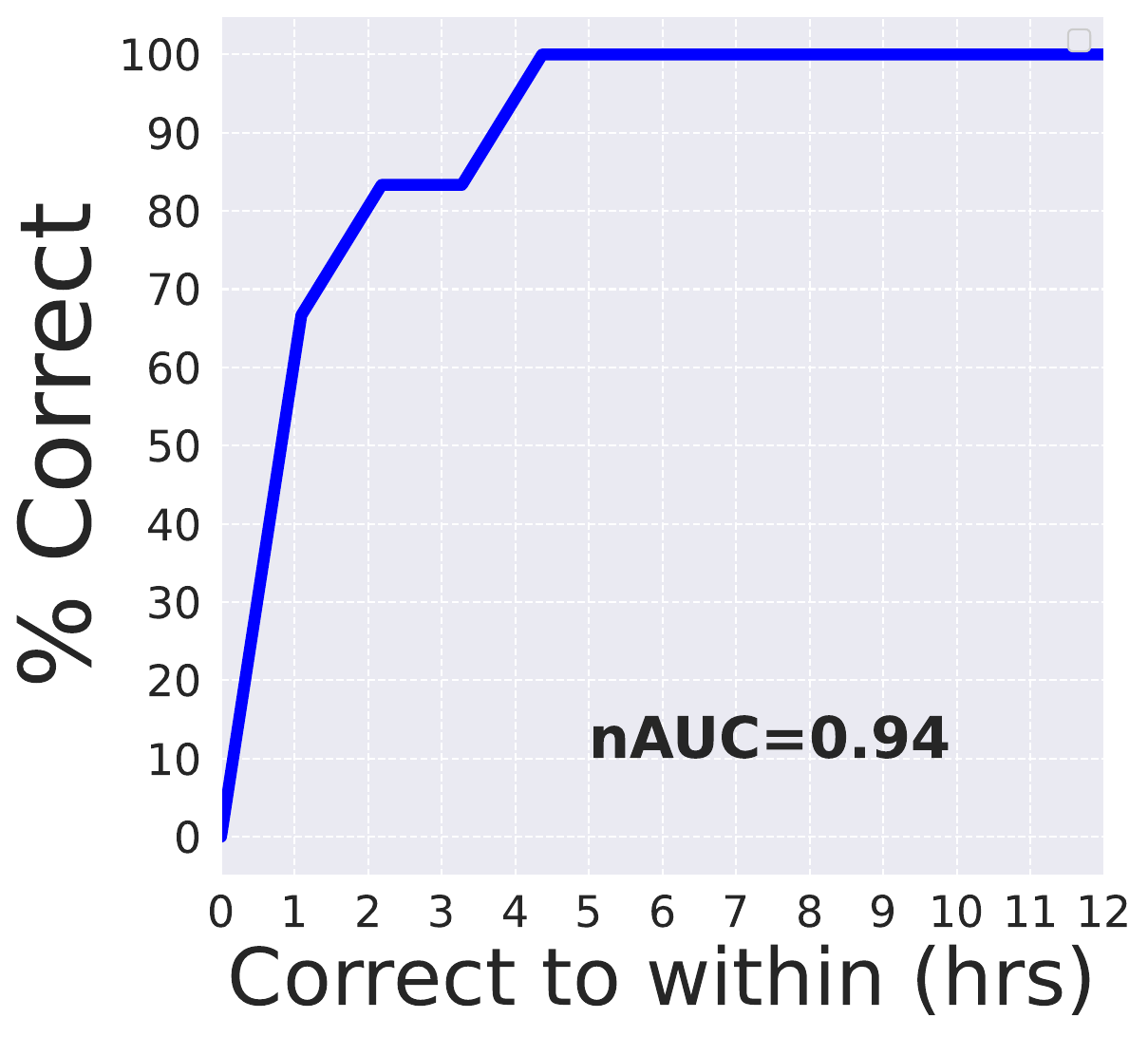} &
\includegraphics[width=.22\textwidth]{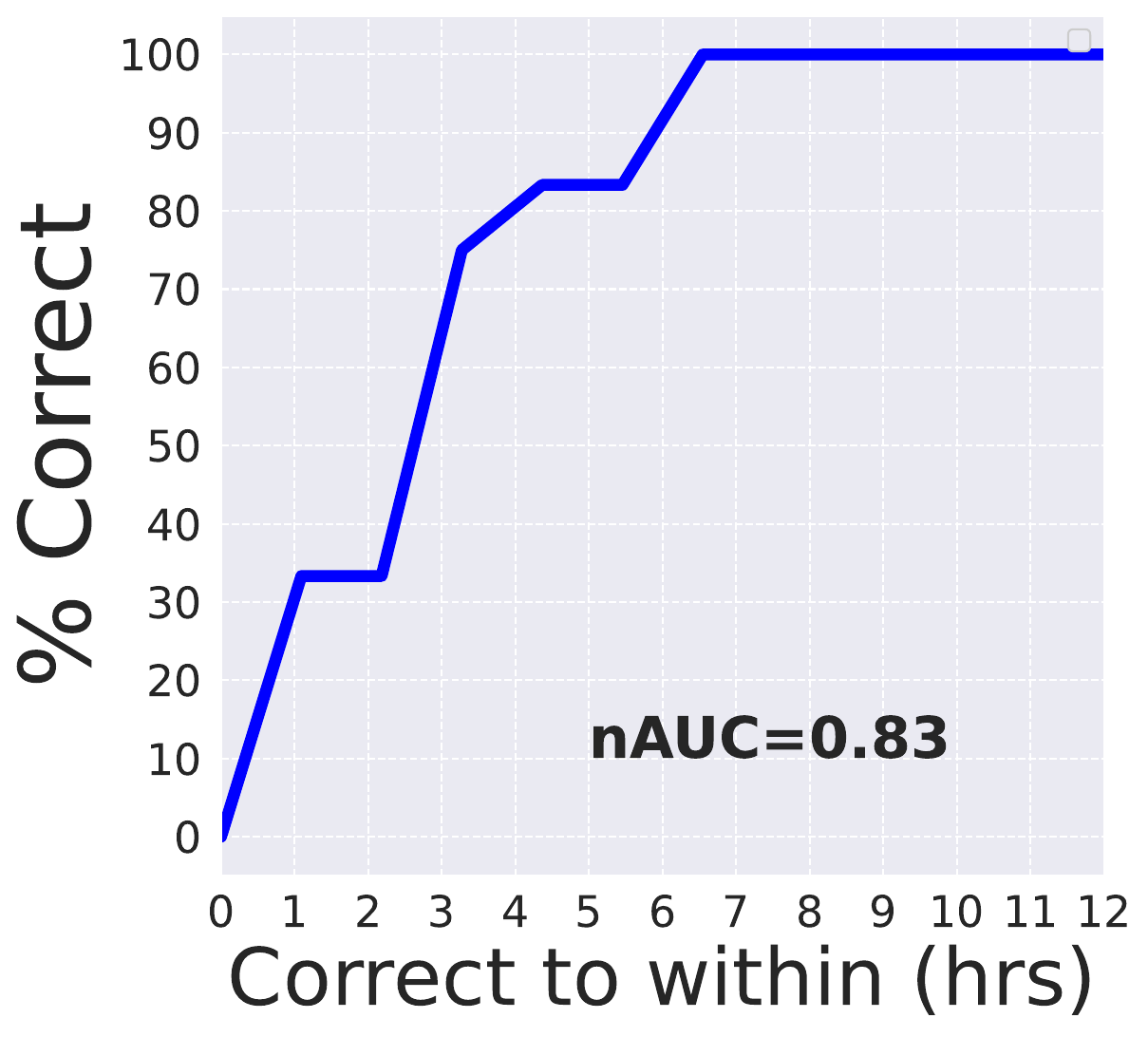}&
\includegraphics[width=.22\textwidth]{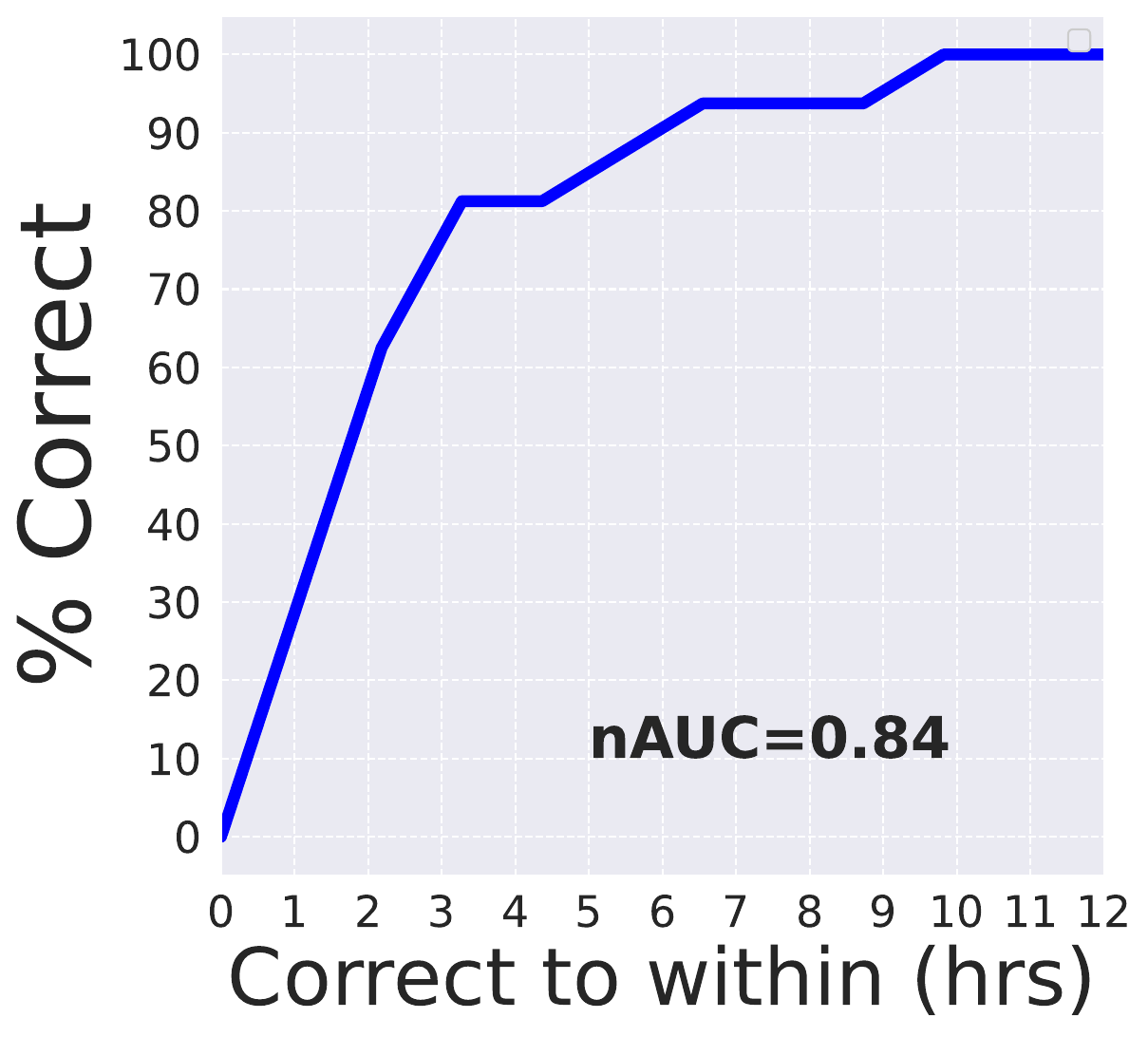}    &
\includegraphics[width=.22\textwidth]{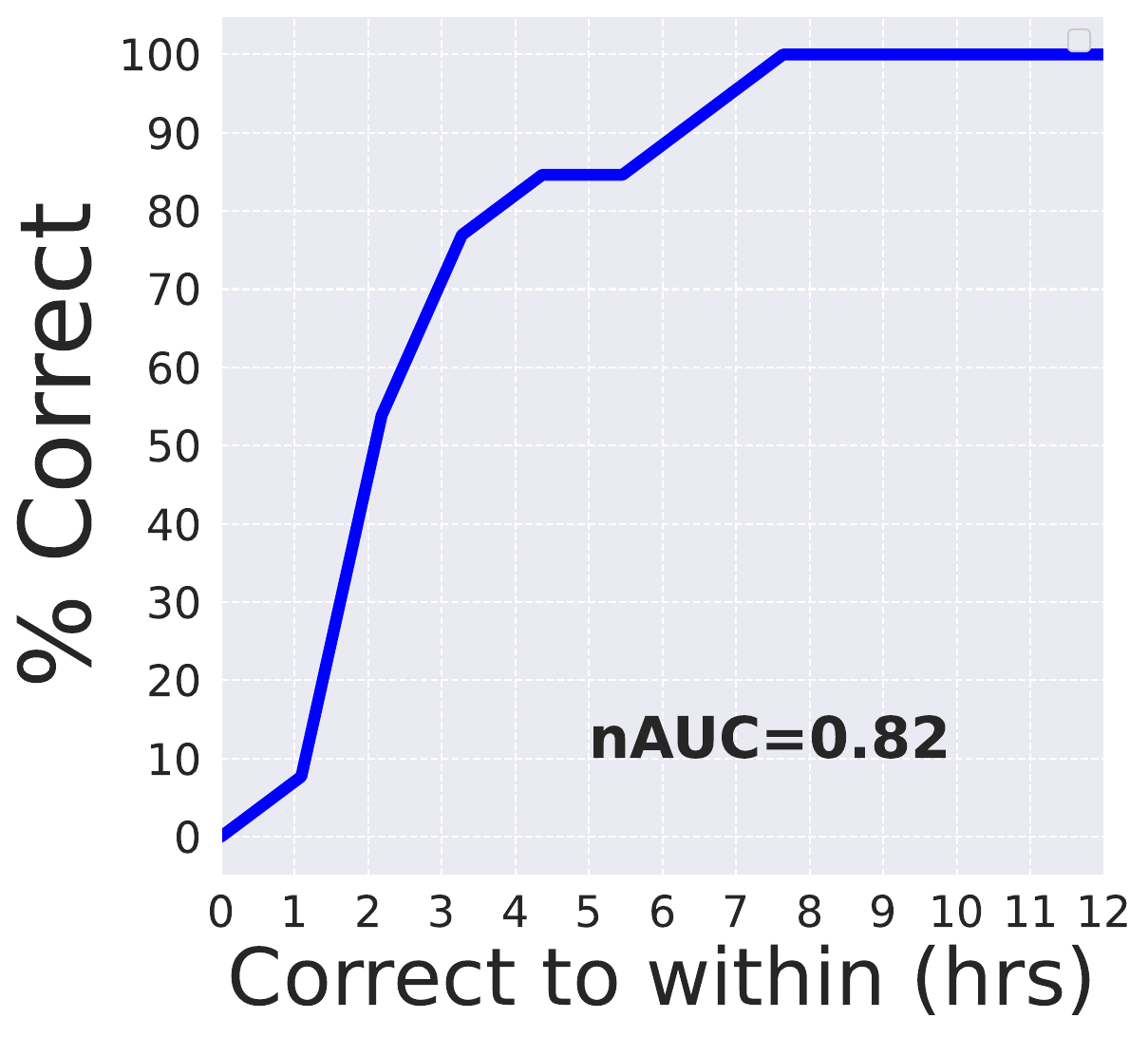} \\

\includegraphics[width=.22\textwidth]{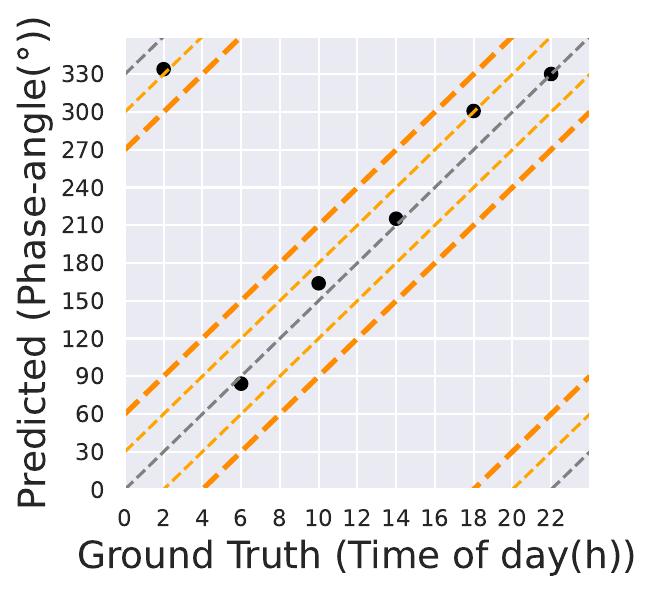} &
\includegraphics[width=.22\textwidth]{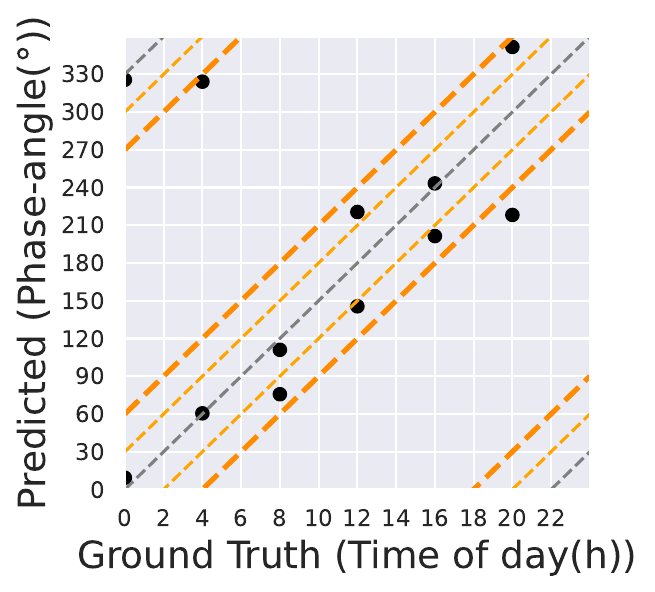}&
\includegraphics[width=.22\textwidth]{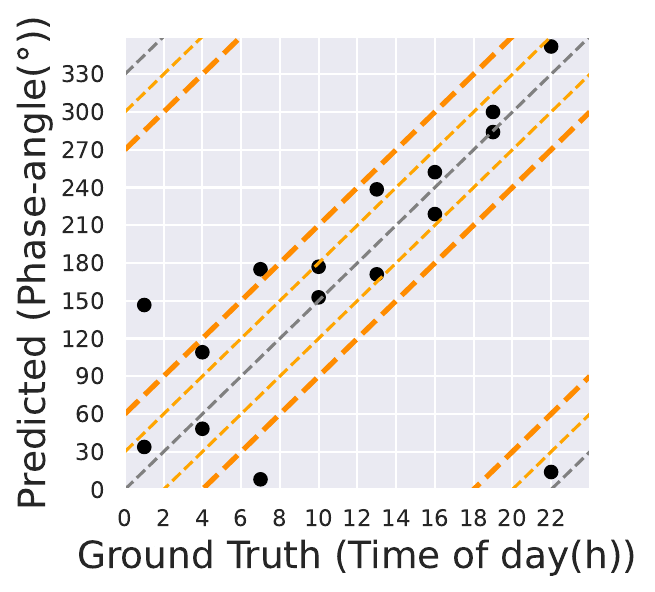} &
\includegraphics[width=.22\textwidth]{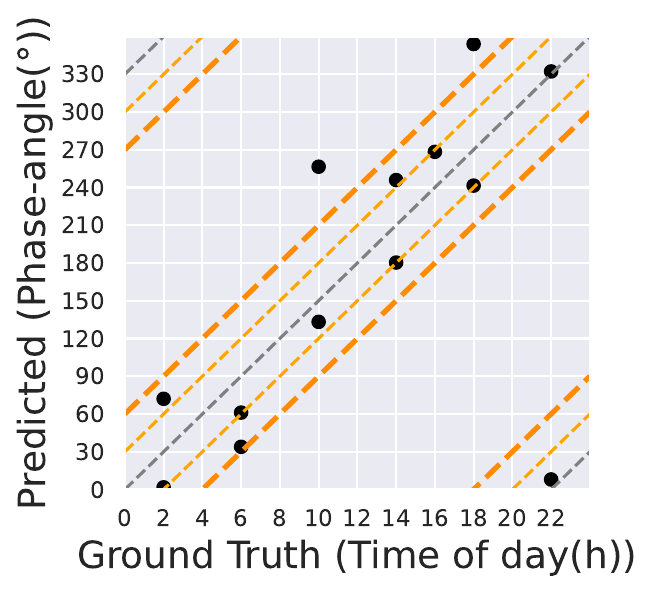} \\
        {(a)} & {(b)}& {(c)} &{(d)} \\

\end{tabular}
\caption{Accuracy of PROTECT on (a) Ostreococcus tauri, (b) Mouse hip articular cartilage, (c) Mouse liver, and (d) Human plasma. The top row shows ROC curves where the y-axis shows the fraction of correctly predicted samples, and the x-axis shows the size of errors. The bottom row shows the scatter plots of predictions vs ground truth.}
\label{fig:ROC}
\end{figure*}

\begin{figure*}[h]
\centering
\begin{tabular}{ccc}
\multicolumn{2}{c}{{CYCLOPS}} & {Ours} \\ 
\includegraphics[width=.30\textwidth]{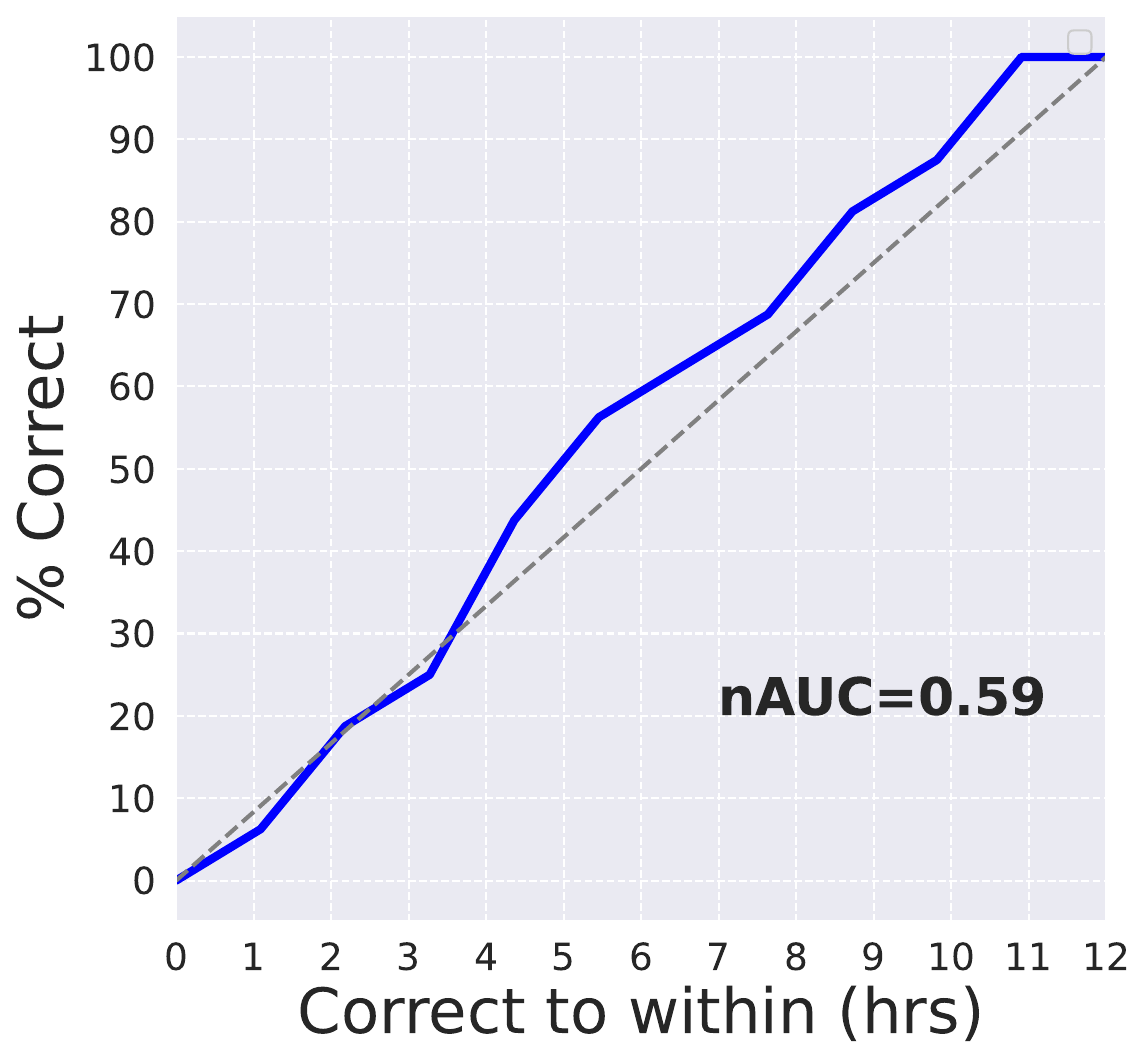} &
\includegraphics[width=.3\textwidth]{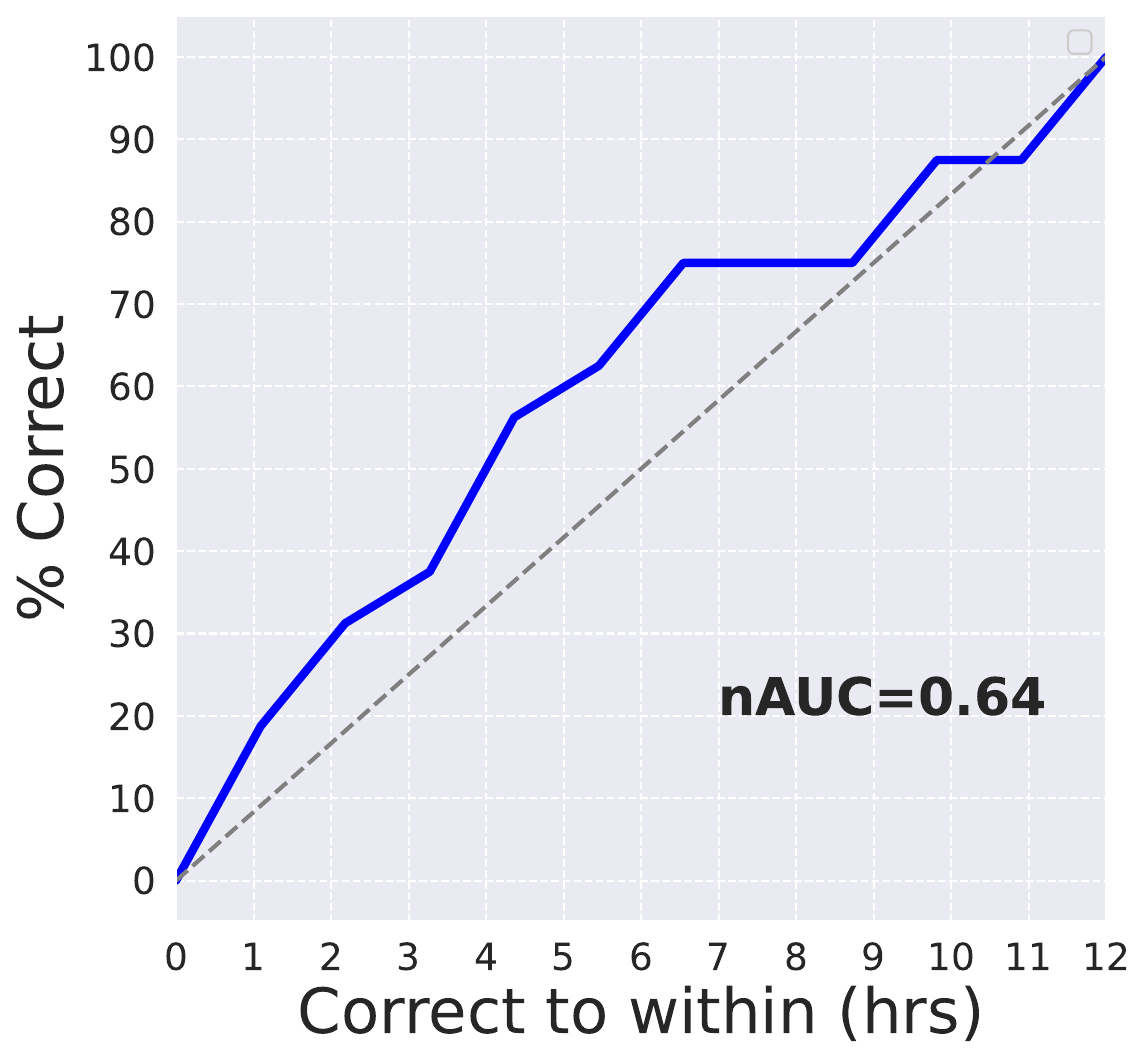}    &
\includegraphics[width=.3\textwidth]{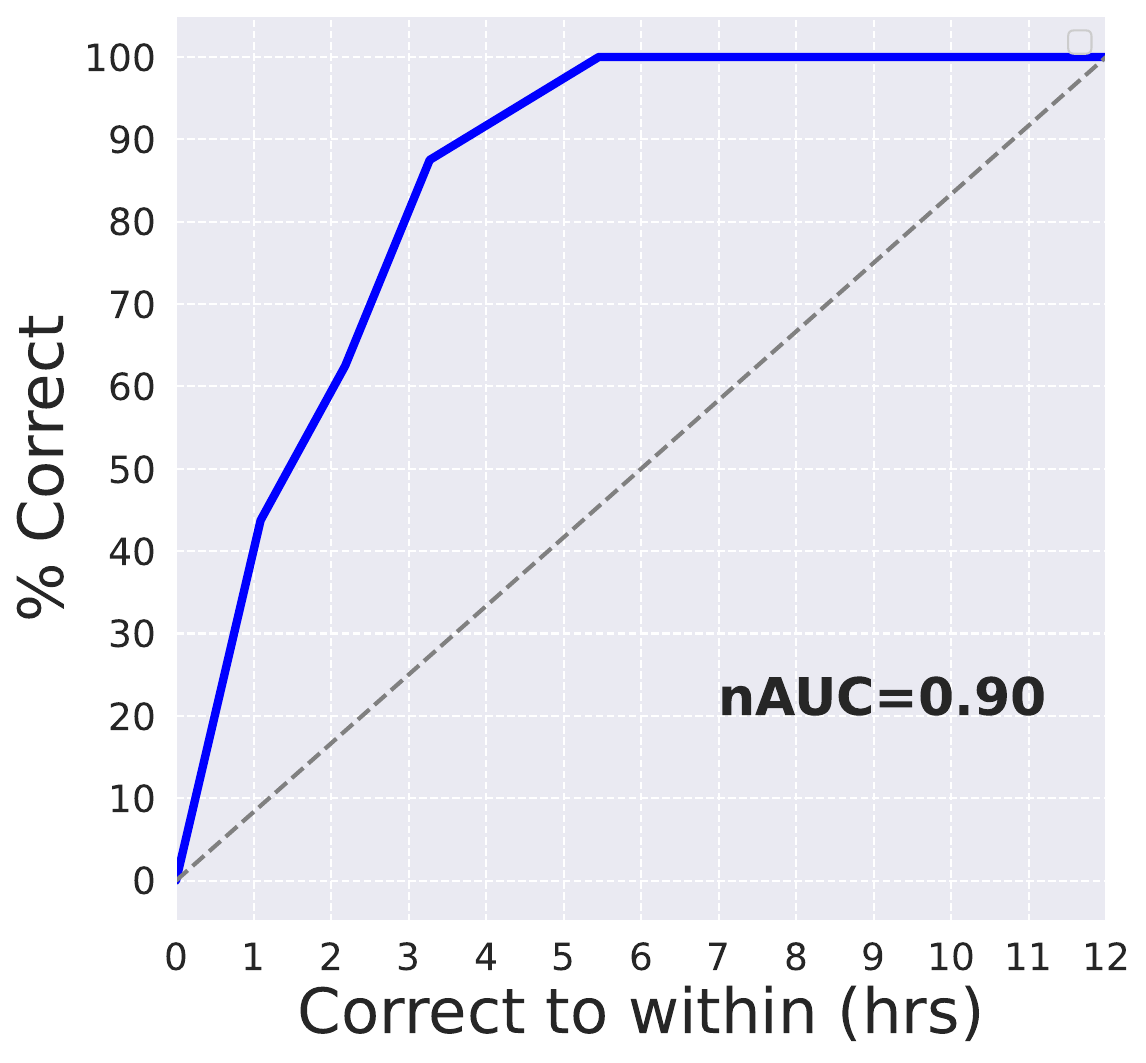} \\
\includegraphics[width=.3\textwidth]{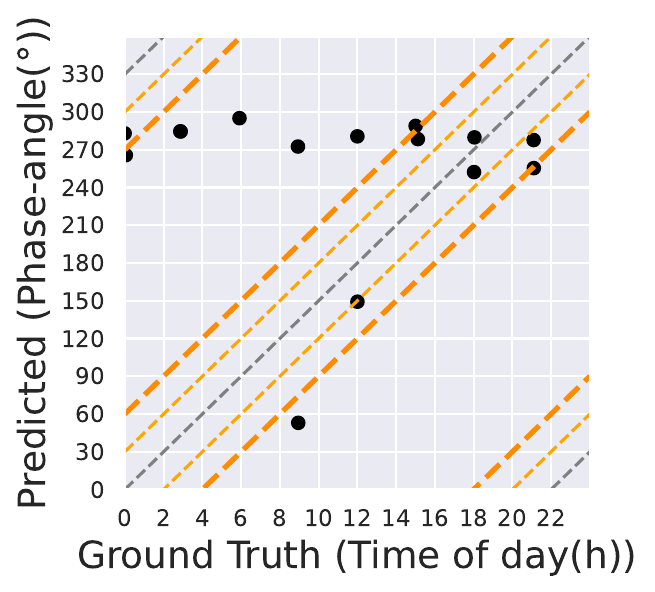} &
\includegraphics[width=.3\textwidth]{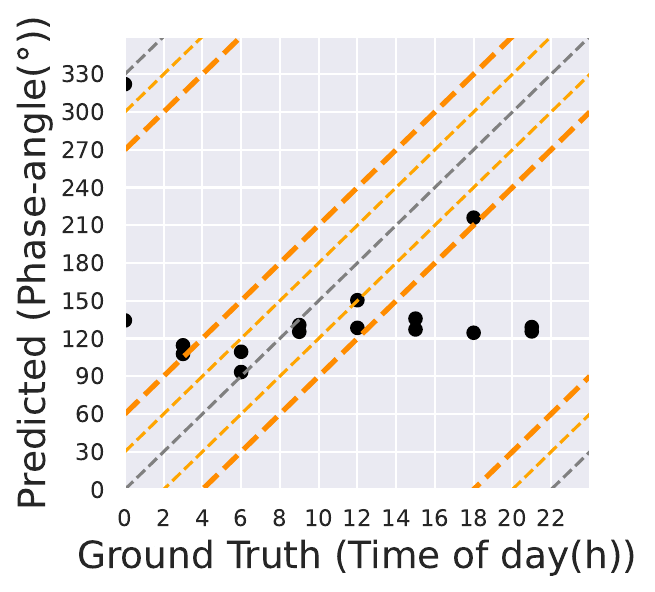}&
\includegraphics[width=.3\textwidth]{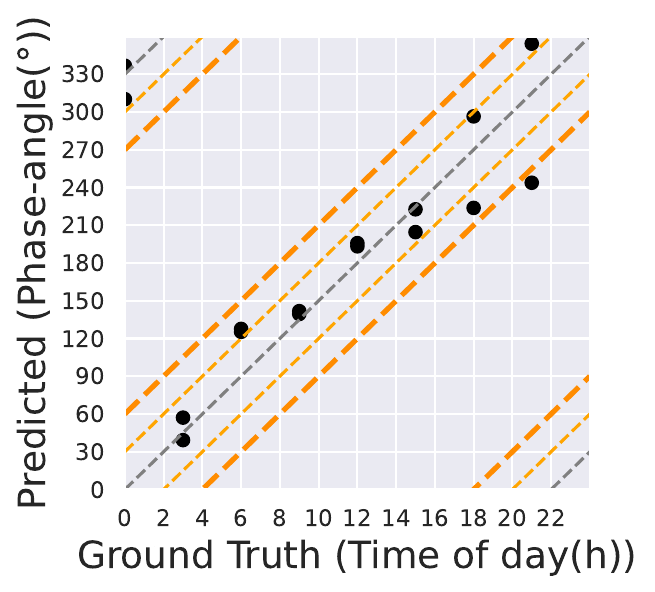} \\
{(a)} & {(b)}& {(c)}  \\
\end{tabular}
\caption{ Comparison of CYCLOPS and our method on mouse liver dataset. (a) CYCLOPS ROC curve and predicted sample phases vs ground truths without using proteins corresponding to seed genes. (b) CYCLOPS ROC curve and predicted sample phases vs ground truths  after using proteins corresponding to seed genes. (c) our  ROC curve and predicted sample phases vs ground truths.}
\label{fig:cyc}
\end{figure*}

\subsection{Labeled proteomic data experiments}
\subsubsection{Data description}
Several proteomic datasets were utilized, sourced from mouse hip articular cartilage (PXD019431) \cite{dudek2021circadian}, two different mouse liver datasets \cite{mauvoisin2014circadian,wang2017nuclear}, Ostreococcus tauri cells \cite{noordally2023phospho} and human plasma \cite{depner2018mistimed}. Mouse hip articular cartilage was collected every four hours over two days, sampling 6 animals at each time point. We calculated the average of label-free quantification (LFQ) intensity across all six animals for each time point, resulting in a set of 12 samples. Both mouse liver datasets consist of 16 samples obtained from mice at 3-hour intervals over a 2-day span. These data quantify the relative protein abundance in each of the 16 samples against a common reference sample, which was labeled using the SILAC method. The dataset mentioned in Wang et al. (2017) \cite{wang2017nuclear} contains about 1000 more proteins than the one referenced in Mauvoisin et al. (2014) \cite{mauvoisin2014circadian}. The Ostreococcus tauri cell dataset provides the mean of normalized abundance per time point, encompassing 6 samples. The human plasma data consists of samples from 6 healthy young males at different time points over two days. We utilized the time points with more than one subject and averaged them. This resulted in time points of 1, 5, 9, 13, 15, 17, and 21 for one day, and 1, 5, 9, 13, 17, and 21 for the second day.

\begin{figure*}
\centering
\begin{minipage}{0.35\textwidth}
\centering
\includegraphics[width=\linewidth,height=0.2\textheight,keepaspectratio]{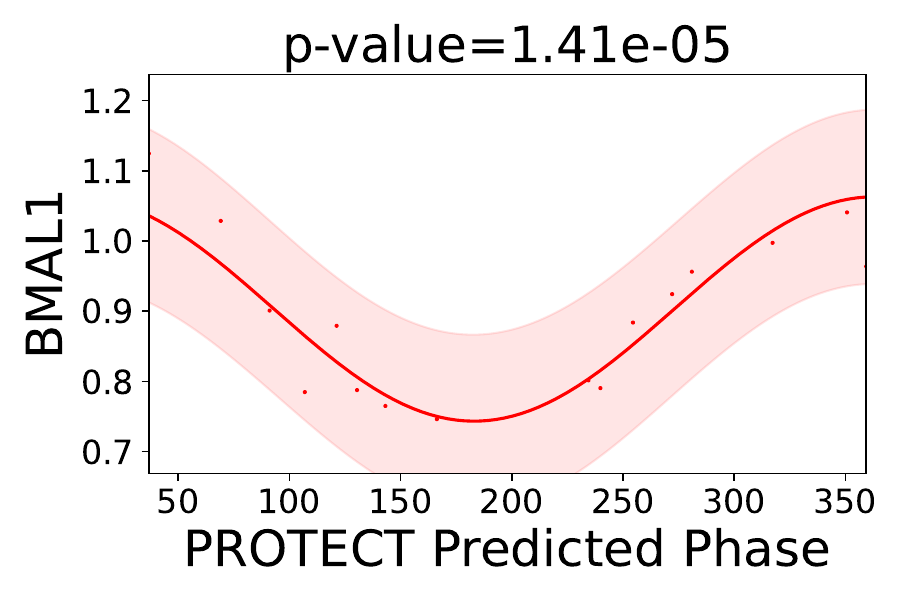}
\label{fig:BMAL1}
\end{minipage}
\begin{minipage}{0.35\textwidth}
\centering
\includegraphics[width=\linewidth,height=0.2\textheight,keepaspectratio]{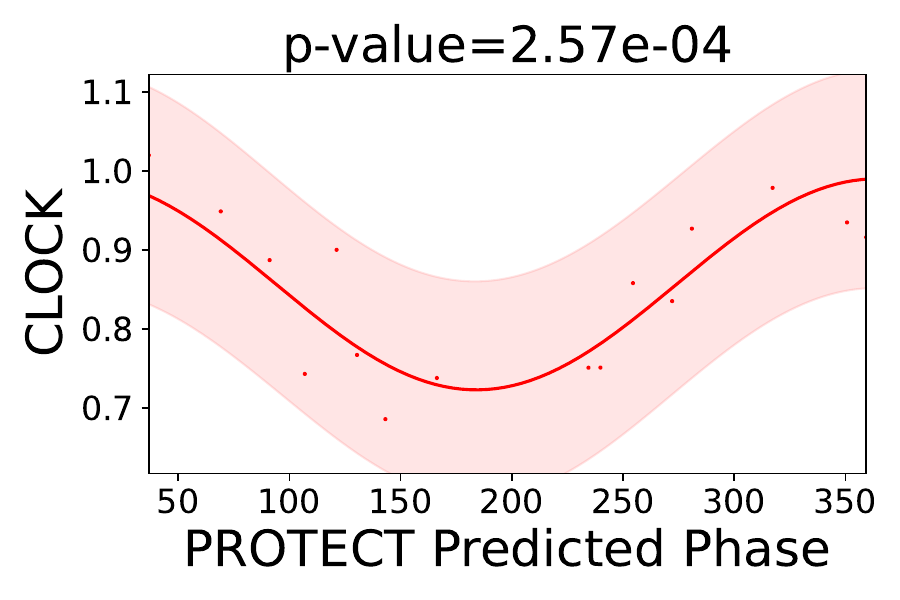}
\label{fig:CLOCK}
\end{minipage}
  
\vspace{0.3cm}
  
\begin{minipage}{0.35\textwidth}
\centering
\includegraphics[width=\linewidth,height=0.2\textheight,keepaspectratio]{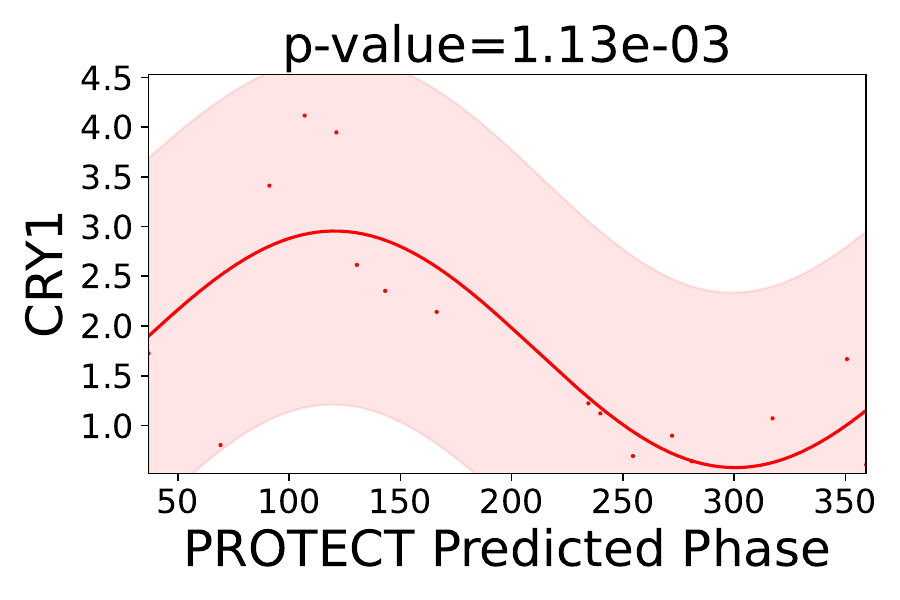}
\label{fig:CRY1}
\end{minipage}
\begin{minipage}{0.35\textwidth}
\centering
\includegraphics[width=\linewidth,height=0.2\textheight,keepaspectratio]{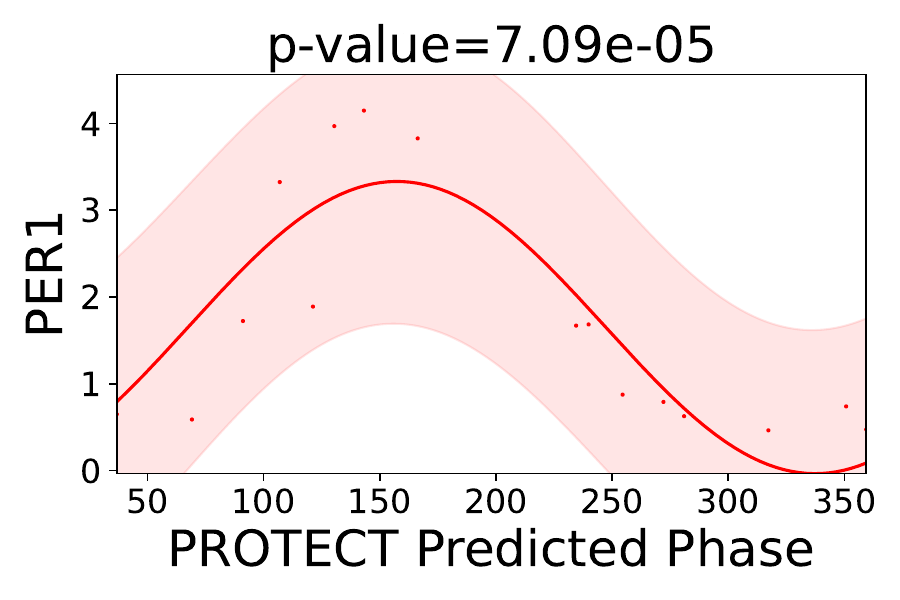}
\label{fig:PER1}
\end{minipage}
  
\caption{Plots of four core clock proteins in mouse liver using predicted phases by PROTECT. The y-axis represents protein expression levels, and the x-axis represents the predicted phases (in degrees) as determined by PROTECT.}
\label{fig:coreclockproteins}
\end{figure*}

\begin{figure*}[h]
\centering
\begin{minipage}{0.35\textwidth}
\centering
\includegraphics[width=\linewidth,height=0.2\textheight,keepaspectratio]{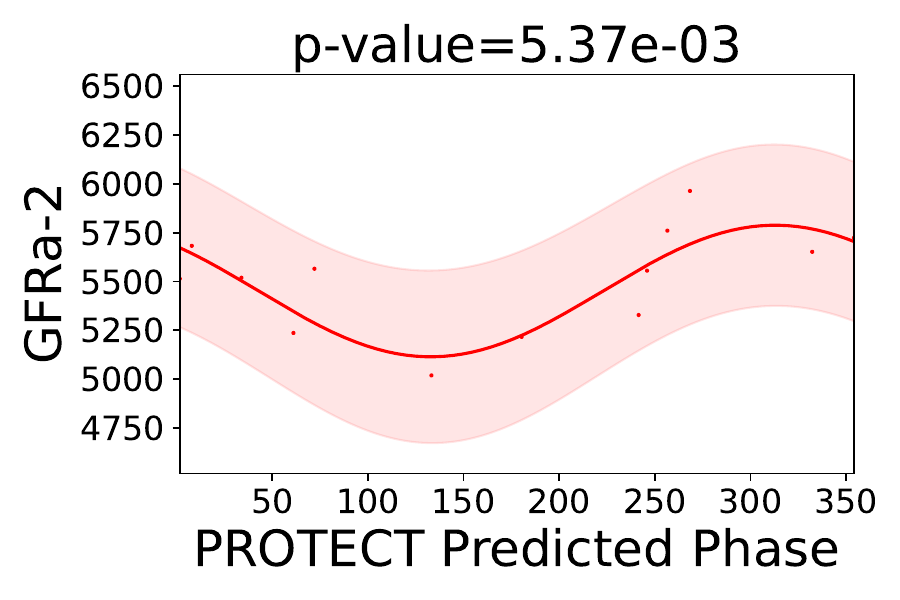}
\label{fig:GFRa-2}
\end{minipage}
\begin{minipage}{0.35\textwidth}
\centering
\includegraphics[width=\linewidth,height=0.2\textheight,keepaspectratio]{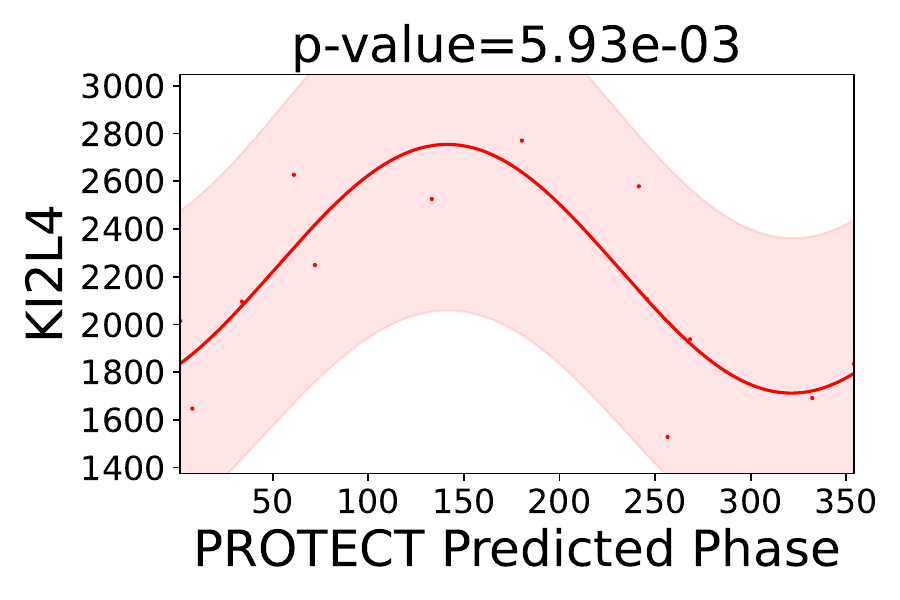}
\label{fig:KI2L4}
\end{minipage}

\vspace{0.3cm}
  
\begin{minipage}{0.35\textwidth}
\centering
\includegraphics[width=\linewidth,height=0.2\textheight,keepaspectratio]{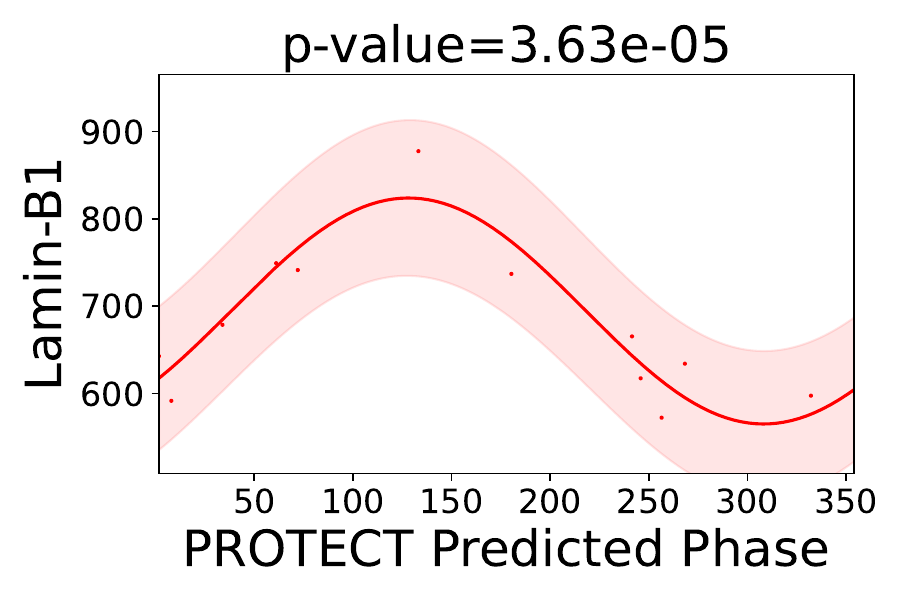}
\label{fig:Lamin-B1}
\end{minipage}
\begin{minipage}{0.35\textwidth}
\centering
\includegraphics[width=\linewidth,height=0.2\textheight,keepaspectratio]{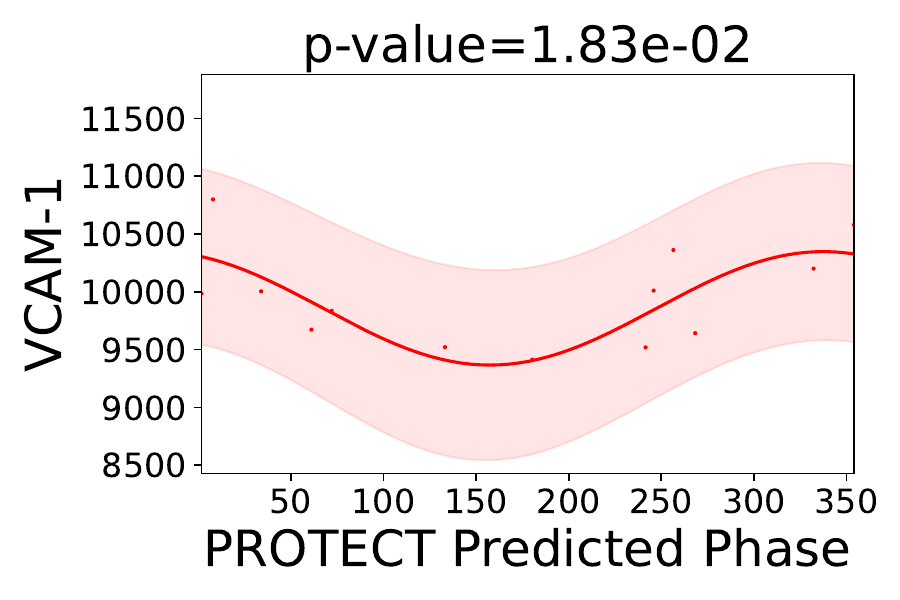}
\label{fig:VCAM-1}
\end{minipage}

\caption{Plots of four randomly chosen proteins known to be strongly regulated by circadian cycle in human plasma using predicted phases by PROTECT. The y-axis represents protein expression levels, and the x-axis represents the predicted phases (in degrees) as determined by PROTECT.}
\label{fig:humancircadian}
\end{figure*}

\subsubsection{Performance evaluation}
 We first demonstrate the time of day \color{black} prediction accuracy of PROTECT on time-labeled datasets, employing ROC curves that show the fraction of correctly predicted sample phases relative to the size of errors \cite{braun2018universal}. Moreover, we show scatter plots of predicted sample phases with respect to the ground truths. 

 PROTECT shows excellent performance on various datasets, achieving a high normalized area under the curve (nAUC) of 94\% on Ostreococcus tauri cell data (Figure \ref{fig:ROC}(a)), and over 80\% on mouse hip articular cartilage (Figure \ref{fig:ROC}(b)), mouse liver \cite{mauvoisin2014circadian} (Figure \ref{fig:ROC}(c)), and human plasma (Figure \ref{fig:ROC}(d)) proteomic datasets. The bottom row of Figure \ref{fig:ROC} demonstrates accurate phase predictions compared to ground truth across all datasets. In particular, all samples of the Ostreococcus tauri cell data show minimal phase prediction errors, while samples in other datasets typically deviate by no more than 4 hours (60 degrees in circadian phase). 
 
We further evaluated the performance of PROTECT on an additional mouse liver dataset \cite{robles2014vivo} and mouse brown adipose tissue (BAT) tissue \cite{qian2023multitissue} with the corresponding results presented in the supplementary material (Figure S6). The results further demonstrate PROTECT’s ability to accurately predict sample phases. \color{black}
 
We tested existing methods, originally designed for gene expression data, on the proteomic mouse liver dataset (PXD003818) \cite{wang2017nuclear}, which includes 16 samples and 5,301 proteins. This dataset was used because it contains the most proteins corresponding to seed rhythmic genes compared to other datasets.
Figure \ref{fig:cyc}(a) shows the ROC curve and the sample phase estimations compared to the time labels using CYCLOPS \cite{anafi2017cyclops} without utilizing proteins corresponding to the seed rhythmic genes. This plot shows that without using proteins corresponding to seed genes, the prediction is close to random guessing with an nAUC of 59\%. Moreover, correct estimates should ideally be on the dotted diagonal gray lines (where lines in the corners account for circadian periodicity). However, most samples are predicted to have similar values. Using the available 1,076 proteins corresponding to the CYCLOPS-designated 8,504 seed rhythmic genes, the ROC curve and scatter plot in Figure \ref{fig:cyc}(b) show that most sample phase predictions still have similar values, with the nAUC improving by only 5\%, indicating CYCLOPS's failure on proteomic data. CIRCUST failed to execute due to the lack of core clock genes for their ``synchronization procedure,'' as pointed out by a CIRCUST author via GitHub discussions. In contrast, our proposed approach (PROTECT) in Figure \ref{fig:cyc}(c) shows high accuracy, with a high nAUC of 90\% on this mouse liver dataset.  

Moreover, we used this mouse liver data to evaluate the robustness of PROTECT when working with fewer samples. We randomly subsampled the data by removing 3, 6, 9, and 12 samples. This process was repeated 4 times for each subsampling scenario. PROTECT was then applied to predict the sample phases, and the average results across the 4 repetitions are shown in Figure S7.  While the AUC of predictions decreases slightly compared to using the full dataset, it remains consistently above 80\% across all sampling levels. This robustness highlights PROTECT's ability to perform reliably, even when part of the data is used, underscoring its utility for sparse high-throughput proteomics data. \color{black}

\subsubsection{Circadian rhythmicity in known rhythmic proteins}
In Figure \ref{fig:coreclockproteins}, we demonstrate the results of plotting four of the core clock proteins' values versus the predicted sample phases using PROTECT across the mouse liver dataset \cite{wang2017nuclear} where the core clock proteins are available. These results highlight the rhythmicity of these proteins using PROTECT's predicted sample phases. Figure \ref{fig:humancircadian} also depicts four proteins, randomly selected from a set of proteins known to be strongly regulated by the circadian cycle in human plasma \cite{depner2018mistimed}. Our predicted sample phases clearly reveal the circadian rhythmic patterns in these proteins as well.
 Additionally, we evaluated PROTECT’s performance on human thyroid tissue \cite{jiang2020quantitative}. To validate the predicted phases, we examined some known rhythmic proteins in thyroid tissue \cite{ruben2018database}. Our results (supplementary Figure S8) confirmed the rhythmicity of these proteins, supported by significant p-values, further demonstrating PROTECT’s reliability in detecting circadian patterns in human samples.
\color{black}

\subsubsection{Outlier \color{black} handling}
 Here, we show the counts of proteins and samples outliers, as defined in Section \ref{sec: outlier}, on our time-labeled datasets. Our algorithm found no outlier samples, suggesting that all residuals for samples are similarly distributed. However, each dataset exhibited a number of outlier proteins, as shown in Table \ref{tab:outlier}. We excluded the outliers from each dataset and retrained the network to handle these protein outliers. 
 We then fitted Cosinor curves against the ground truth and verified that these proteins did not exhibit significant rhythmicity at a p-value threshold of 5e-2.

\begin{table}[h]
\centering 
\begin{tabular}{|l|c|}
\hline
\textbf{Dataset}      & \textbf{Number of outliers}  \\
\hline
Mouse liver (Wang et al.)                  &          0       \\
Mouse hip articular cartilage            &    13     \\
Mouse liver (Mauvoisin et al.)           &          42     \\
Ostreococcus tauri                &  15  \\
Human plasma           &  21\\
\hline
\end{tabular}
\caption{Number of protein outliers detected in each analyzed dataset.}
\label{tab:outlier}
\end{table}

\subsection{Un-labeled proteomic data experiments}
\subsubsection{Data description}
In our exploration of circadian rhythms in unlabeld human  proteomic data, we examined three brain regions: the Temporal Cortex (TC) \cite{johnson2020large}, the Parietal Association Cortex \cite{carlyle2021synaptic}, and the Dorsolateral Prefrontal Cortex (DLPFC) \cite{beach2015arizona,johnson2020large}. Additionally, we incorporated a dataset derived from urine samples \cite{wang2023identification}. In the TC and DLPFC datasets, the LFQ intensity is used for protein quantitation. In the parietal association cortex dataset, the calculation of protein intensities involves summing the TMT reporter ions corresponding to all peptides assigned to each protein. In the urine dataset, protein quantification is conducted utilizing the intensity-based absolute quantification (iBAQ) algorithm.

\subsubsection{Data preparation for control and AD subjects comparison}
In preparing data for comparing control and AD subjects in unlabeled brain datasets, we took several steps. Initially, we removed proteins with missing values from both control and AD datasets. To ensure a fair comparison, we then used the common set of proteins in both groups. Moreover, to handle different sample sizes in control and AD groups, particularly in TC and DLPFC datasets, we made them equal by matching age and sex characteristics. In the urine dataset with numerous missing values, we removed proteins with missing values in over 50\% of samples, following a similar approach as in the original paper \cite{wang2023identification}. Then, we found the intersection of all proteins in control, MCI, and AD subjects.

\subsubsection{Disparities between AD and control subjects}
To find rhythmic proteins for control and AD groups, we used stringent criteria which involved utilizing the Benjamini-Hochberg False Discovery Rate (FDR), relative amplitude (rAmp) (amplitude divided by the baseline), and coefficient of determination ($R^2$).  
We utilized CosinorPy \cite{movskon2020cosinorpy} to fit each protein with a cosine curve, employing our predicted phases to determine these values for each protein. Subsequently, we applied predetermined thresholds for significance as follows: FDR $< 0.05$, rAmp $\geq 0.1$, and $R^{2} \geq 0.1$.  Proteins that met all these criteria were considered rhythmic.

\begin{figure*}[hp]
\centering

\begin{minipage}{0.4\textwidth}
\begin{subfigure}{\textwidth}
\centering
\includegraphics[width=0.9\textwidth]{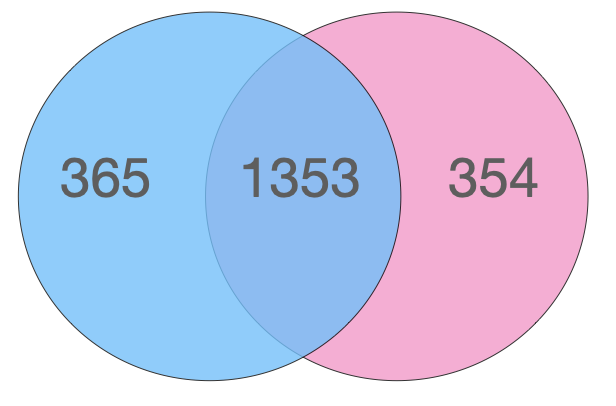}
\caption{}
\label{fig:venn_PC}
\end{subfigure}
    
\vspace{1em} 

\begin{subfigure}{1.3\textwidth}
\centering
\includegraphics[width=1.\textwidth]{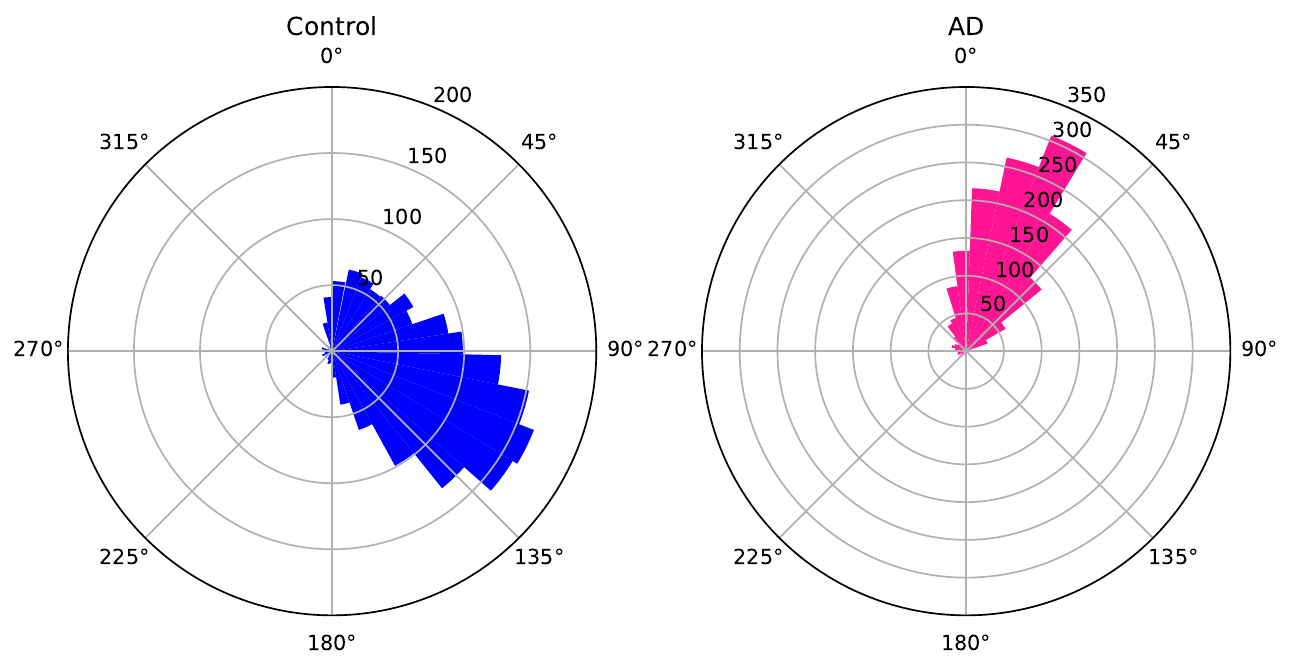}
\caption{}
\label{fig:phases_TC}
\end{subfigure}
\end{minipage}
\hfill
\begin{minipage}{0.5\textwidth}
\centering
\begin{tabular}{cc}
\includegraphics[width=.4\textwidth]{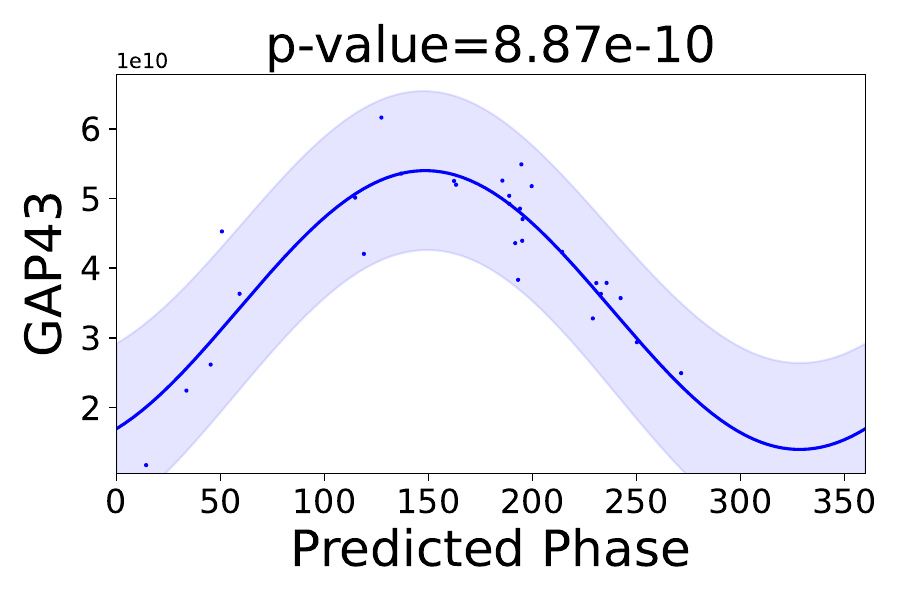} &
\includegraphics[width=.4\textwidth]{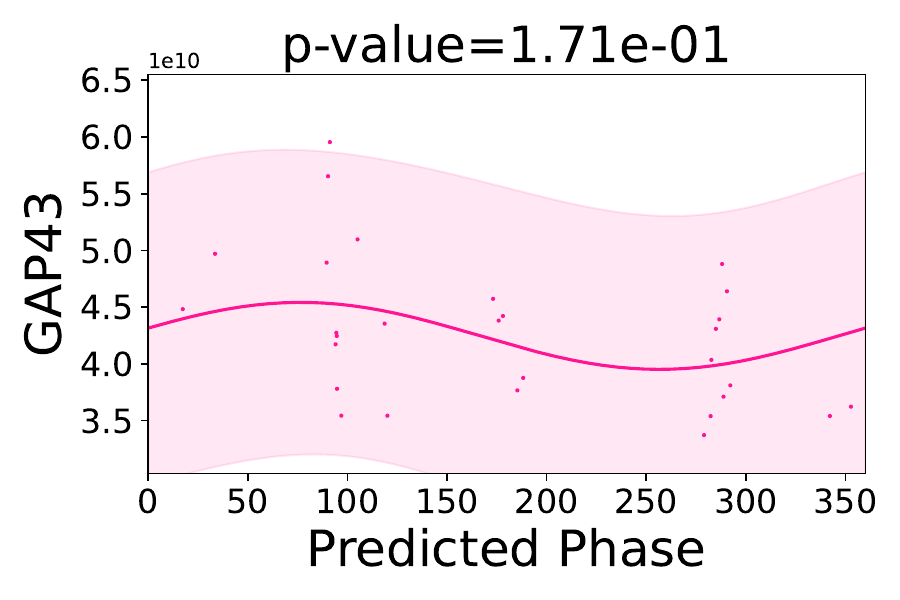} \\
\includegraphics[width=.4\textwidth]{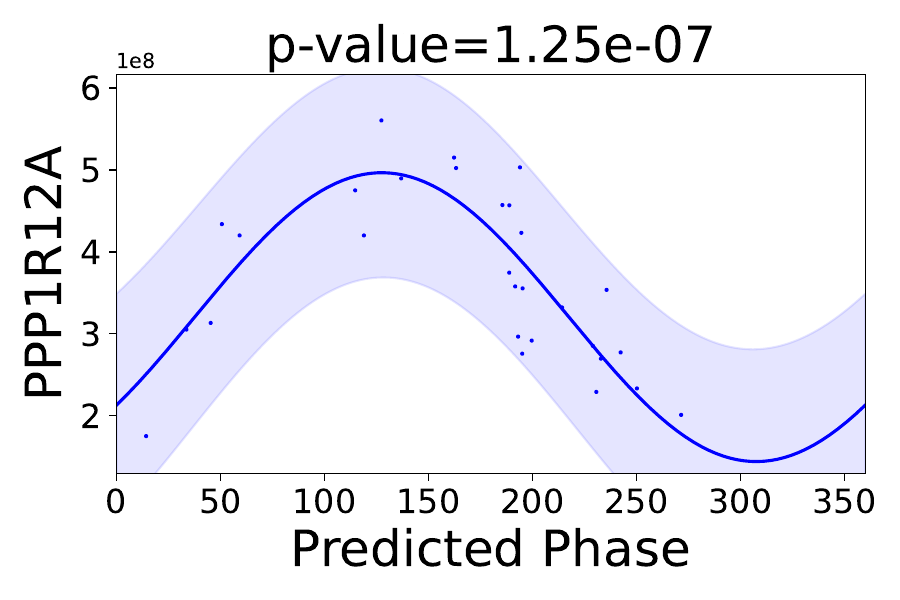} &
\includegraphics[width=.4\textwidth]{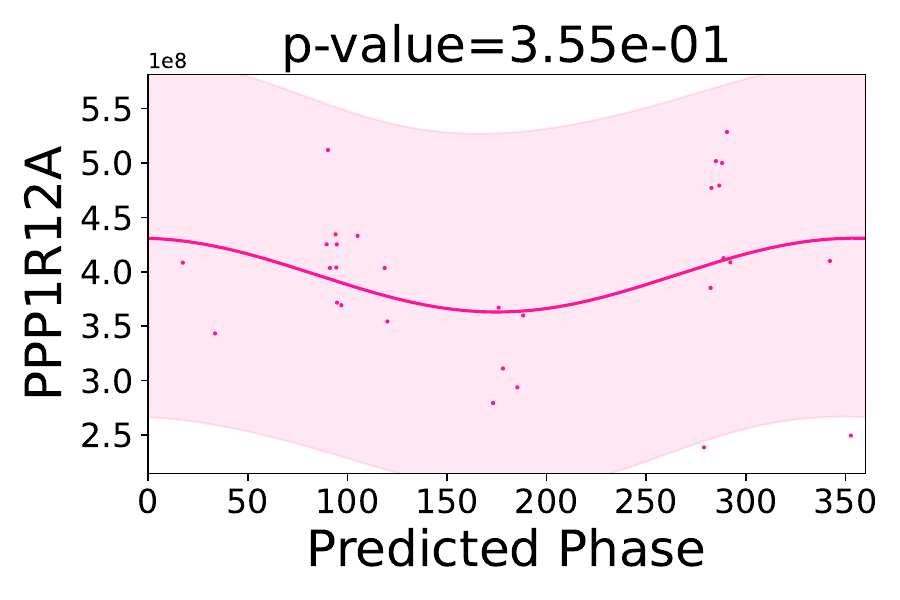} \\
\includegraphics[width=.4\textwidth]{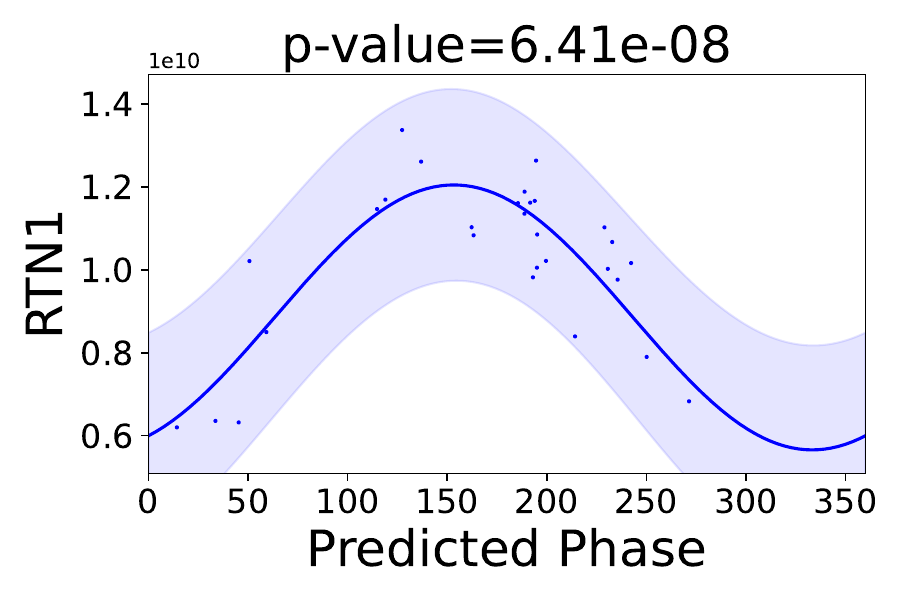} &
\includegraphics[width=.4\textwidth]{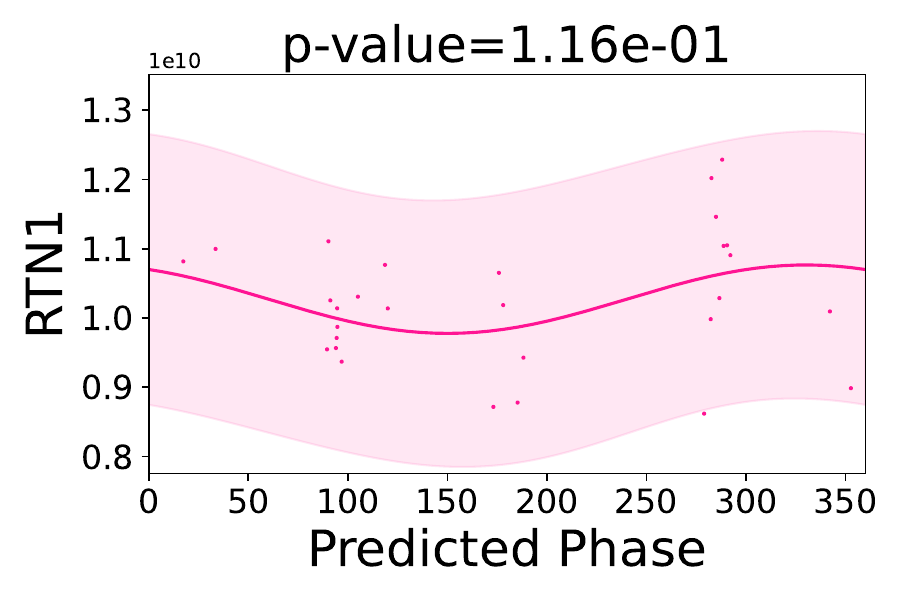} \\
\includegraphics[width=.4\textwidth]{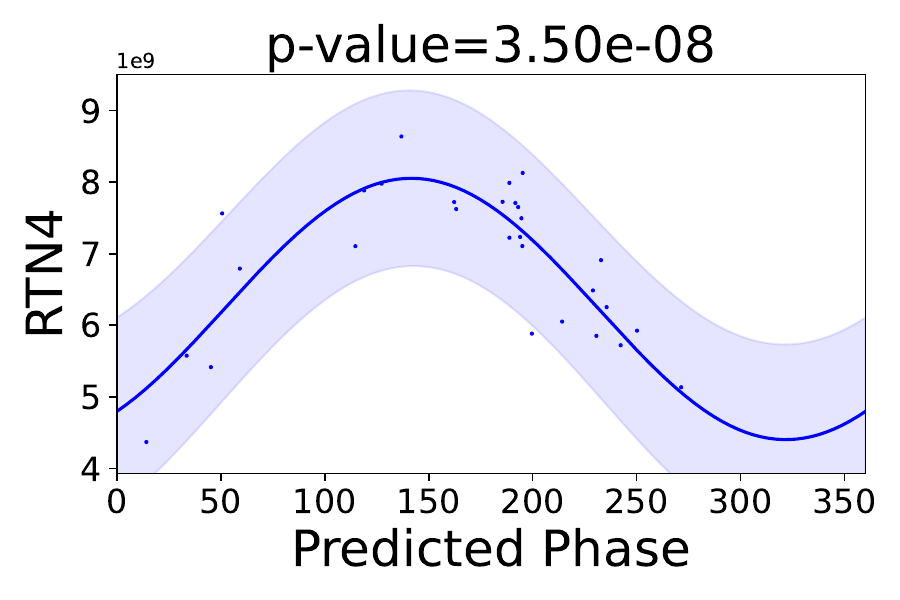} &
\includegraphics[width=.4\textwidth]{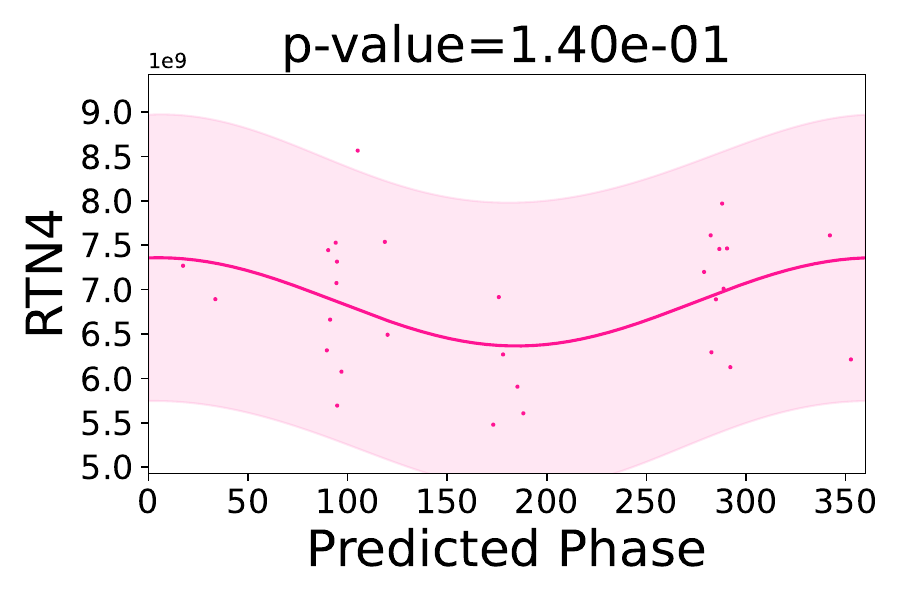} \\
\includegraphics[width=.4\textwidth]{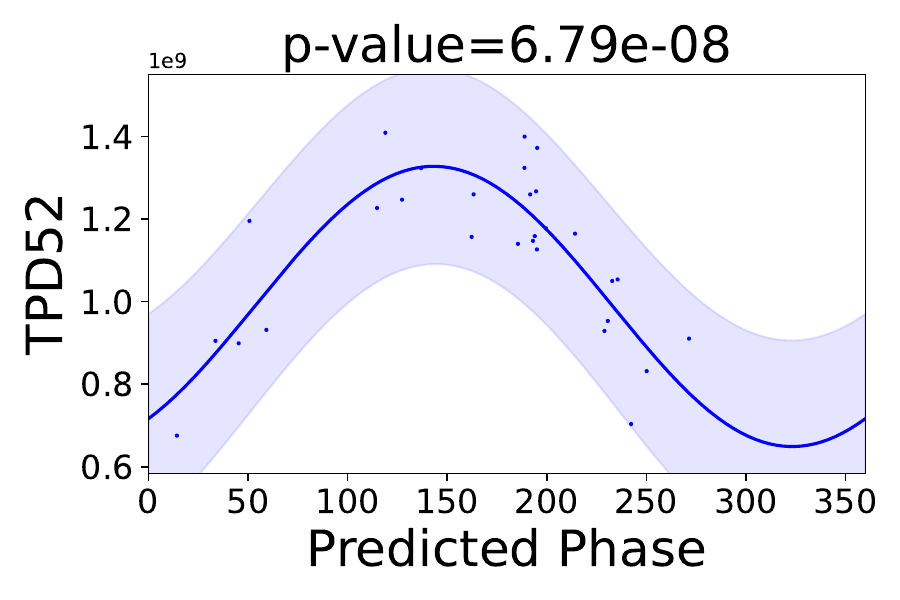} &
\includegraphics[width=.4\textwidth]{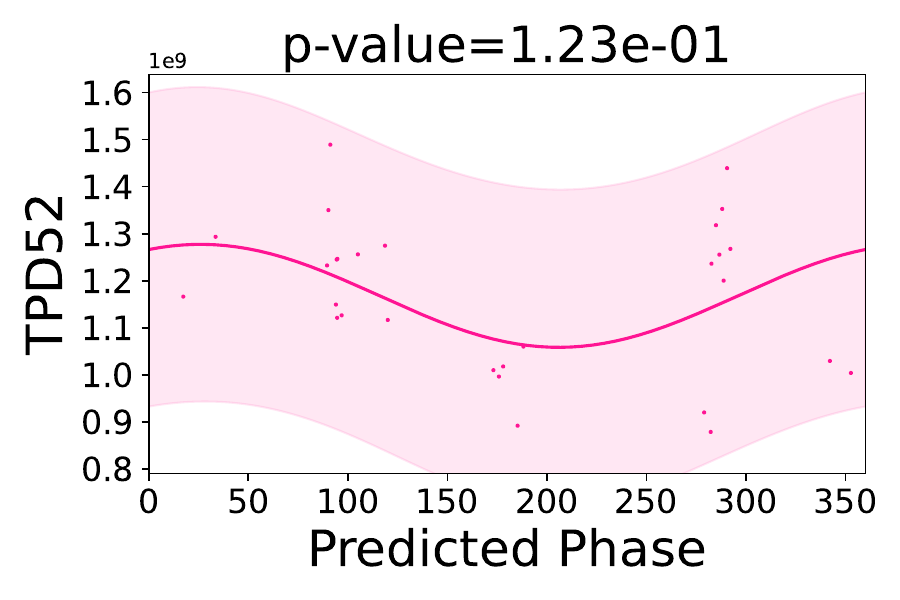} \\
\multicolumn{2}{c}{(c)} \\\\
\end{tabular}
\label{fig:expressions_TC}
\end{minipage}

\caption{Disparities between control and AD subjects in temporal cortex using PROTECT predicted phases. 
(a) Venn diagram of numbers of rhythmic proteins in control (blue) and AD (pink) subjects. 
(b) Rose plots of distributions of peak times (i.e., acrophases) in rhythmic proteins of control and AD subjects within 24 hours (360 degrees) of the circadian cycle. The radial distance indicates protein counts. The max of radial distance differs between control and AD plots. 
(c) Plots of 5 example rhythmic proteins in control subjects that lose rhythmicity in AD.}
\label{fig:TC}
\end{figure*}

Figures \ref{fig:TC}, S9, and S10 illustrate the disparities between control and AD subjects in the TC, parietal association cortex, and DLPFC brain regions. 
First, Venn diagrams are provided to visualize the overlap of rhythmic proteins between control and AD in each brain region (Figures \ref{fig:TC}(a), S9(a), and S10(a)). The results indicate a close similarity in the number of rhythmic proteins between AD and control subjects within each brain region, with a higher count in control subjects. 
 In the TC and DLPFC regions, large proportions of proteins exhibit rhythmicity. In the TC region, out of 2425 proteins, 1718 and 1707 proteins demonstrate rhythmic patterns in control and AD, respectively and 80\% of rhythmic proteins identified in control subjects maintain their rhythmic nature in AD. Also, in the DLPFC region, out of 2483 proteins, 1980 and 1759 proteins display rhythmicity in control and AD, respectively, with 1610 proteins shared between the two groups. However, in the parietal association cortex lower proportions of proteins show rhythmicity. Specifically, out of 3389 proteins, 717 exhibit rhythmicity in control and 692 proteins are rhythmic in AD and 255 rhythmic proteins are shared between the two groups.

In Figures \ref{fig:TC}(b), S9(b), and S10(b), the distribution of peak times of rhythmic proteins in control and AD subjects is depicted within 24 hours (360 degrees) of circadian cycle.  To ensure consistency across datasets, we used EHD1 - a protein that exhibited rhythmicity in all three brain regions and urine samples on different groups (we used Venn diagrams to find overlapping proteins between all different datasets, similar to Figure \ref{fig:TC}(a)) - as a reference point. We set the phase of EHD1 to zero and shifted the acrophases of other proteins relative to this one, thereby providing a common baseline for comparing peak times across regions and conditions. \color{black} Dispersed protein peak distributions arise in the parietal association cortex unlike the TC and DLPFC. In the latter two regions, the peak times of rhythmic proteins are more concentrated in smaller areas and many proteins peak around same times. Additionally, peak times show rotational shifts between control and AD subjects in both the parietal association cortex and TC.  In the TC region, peak times for control subjects are mostly between 90-135 degrees, whereas in AD subjects, they are concentrated between 0-45 degrees. In control subjects within the parietal association cortex region, the majority of peak times occur between 0-45 and 180-225 degrees, whereas in AD subjects, they are observed between 135-180 and 315-360 degrees.

We then illustrate five top proteins that demonstrate significant rhythmicity in control subjects but lose rhythmicity in AD subjects, based on the three criteria mentioned earlier, for each brain region. This is depicted in Figures \ref{fig:TC}(c), S9(c), and S10(c) on TC, parietal association cortex, and DLPFC regions, respectively. Importantly, these proteins not only show significantly increased p-values in AD subjects but also undergo a noticeable reduction in amplitude. 

These findings highlight the differences in the rhythmic protein profiles between control and AD subjects across distinct brain regions. Overall, while the TC and DLPFC show large overlapping circadian proteomic signatures between disease and control states, the parietal association cortex exhibits far fewer shared rhythmic proteins. The variability indicates regionally specific circadian disruptions arise in AD. Identifying the underlying mechanisms driving these cortical differences in rhythmicity may open new therapeutic avenues for future exploration. 

We also investigated the number of rhythmic proteins shared across all three brain regions, categorizing them into four groups: proteins rhythmic in all control subjects across different brain regions, proteins rhythmic in all AD subjects across different brain regions, proteins shared in all controls but not in ADs, and proteins shared in all ADs but not in controls. Table S1 summarizes these findings, revealing 276 proteins exhibits rhythmicity in all control subjects, 271 rhythmic proteins common to all AD subjects, and 4 proteins rhythmic in all controls but not in ADs. No proteins found to be rhythmic in all ADs but not in controls.

\subsubsection{Enrichment analysis}

\begin{figure*}[]
\centering
\begin{subfigure}{0.48\textwidth}
\includegraphics[width=\linewidth, height=0.5\textwidth]{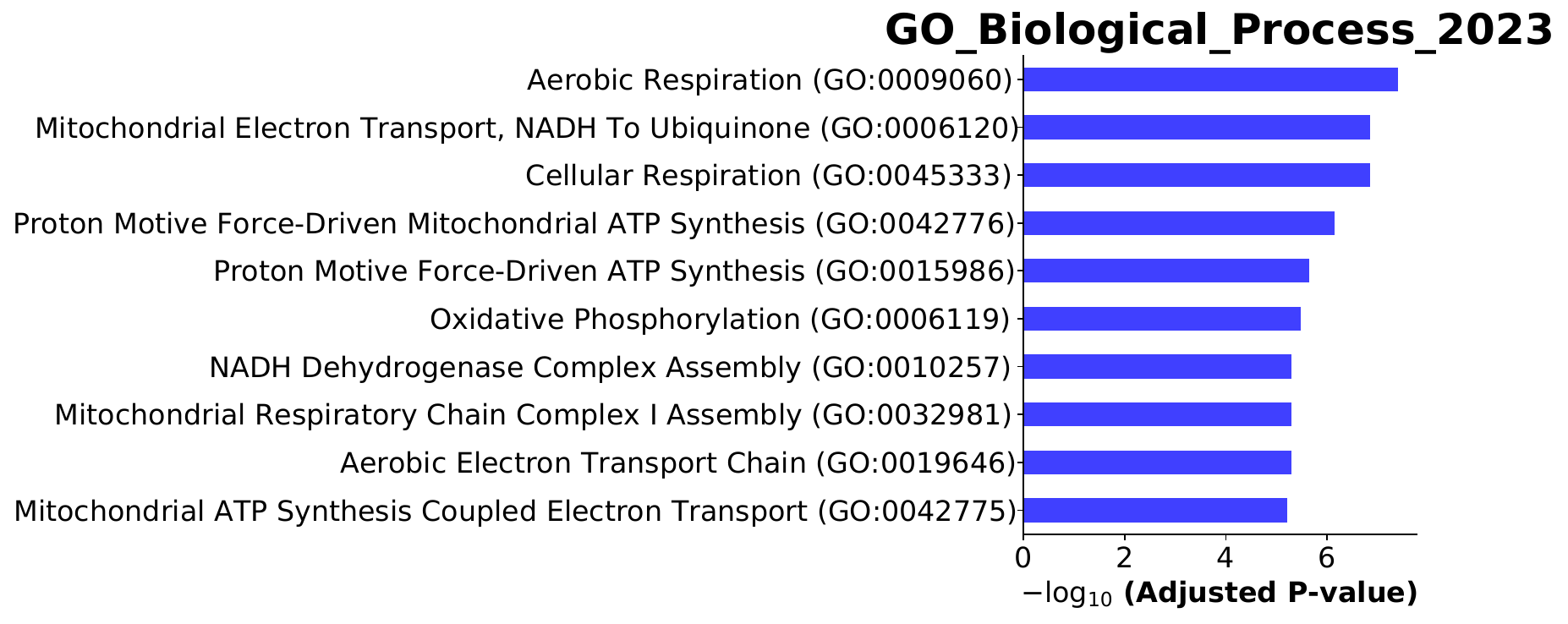}
\end{subfigure}
\hfill
\begin{subfigure}{0.5\textwidth}
\includegraphics[width=\linewidth, height=0.5\textwidth]{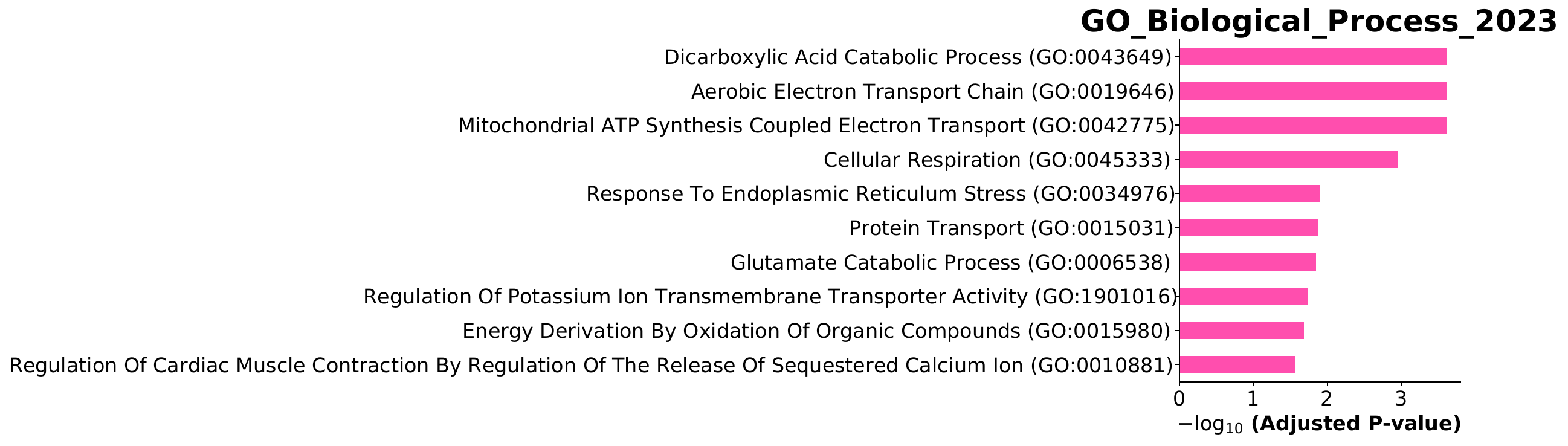}
\end{subfigure}

\vspace{0.5cm}
    
\begin{subfigure}{0.45\textwidth}
\includegraphics[width=\linewidth, height=0.5\textwidth]{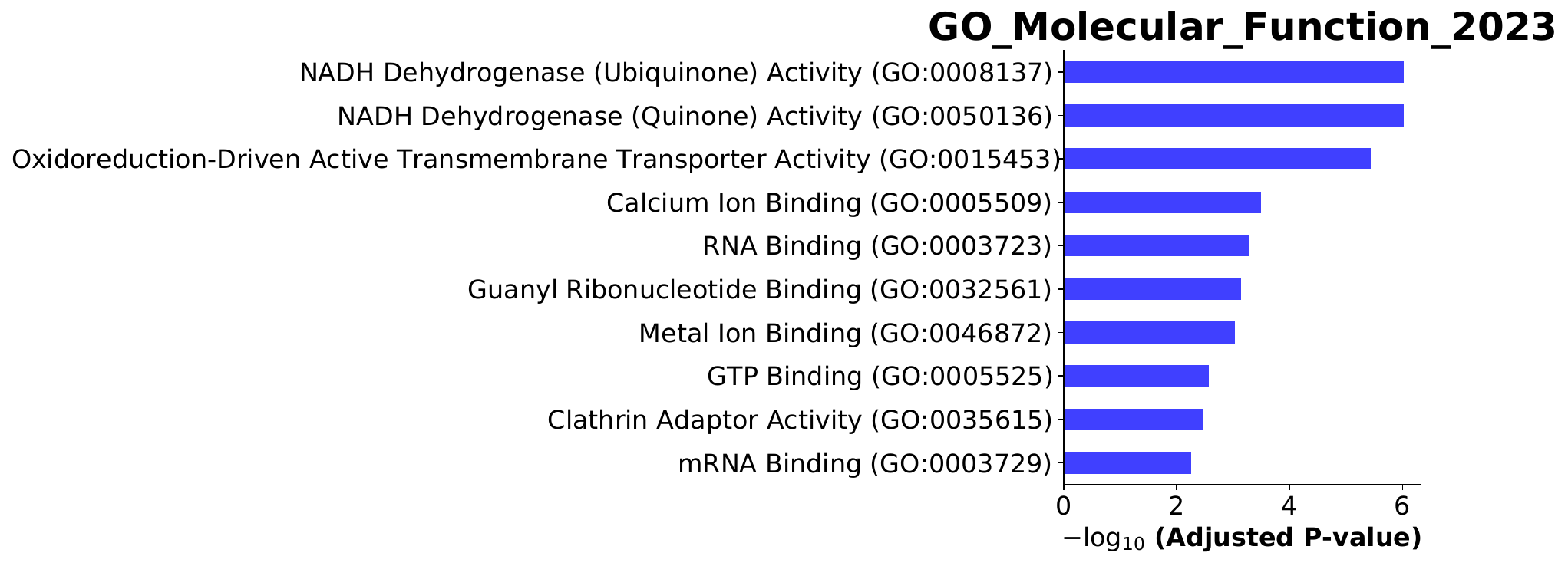}
\end{subfigure}
\hfill
\begin{subfigure}{0.5\textwidth}
\includegraphics[width=\linewidth, height=0.5\textwidth]{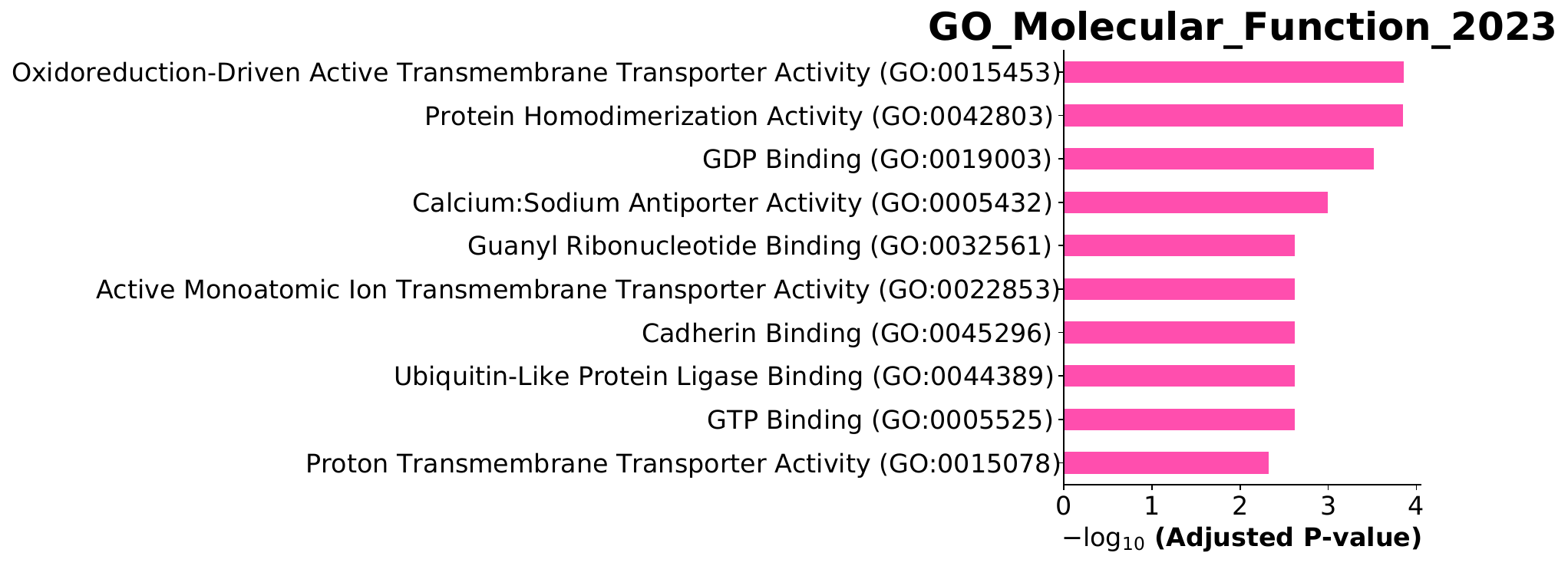}
\end{subfigure}
    
\vspace{0.5cm}
    
\begin{subfigure}{0.45\textwidth}
\includegraphics[width=\linewidth, height=0.5\textwidth]{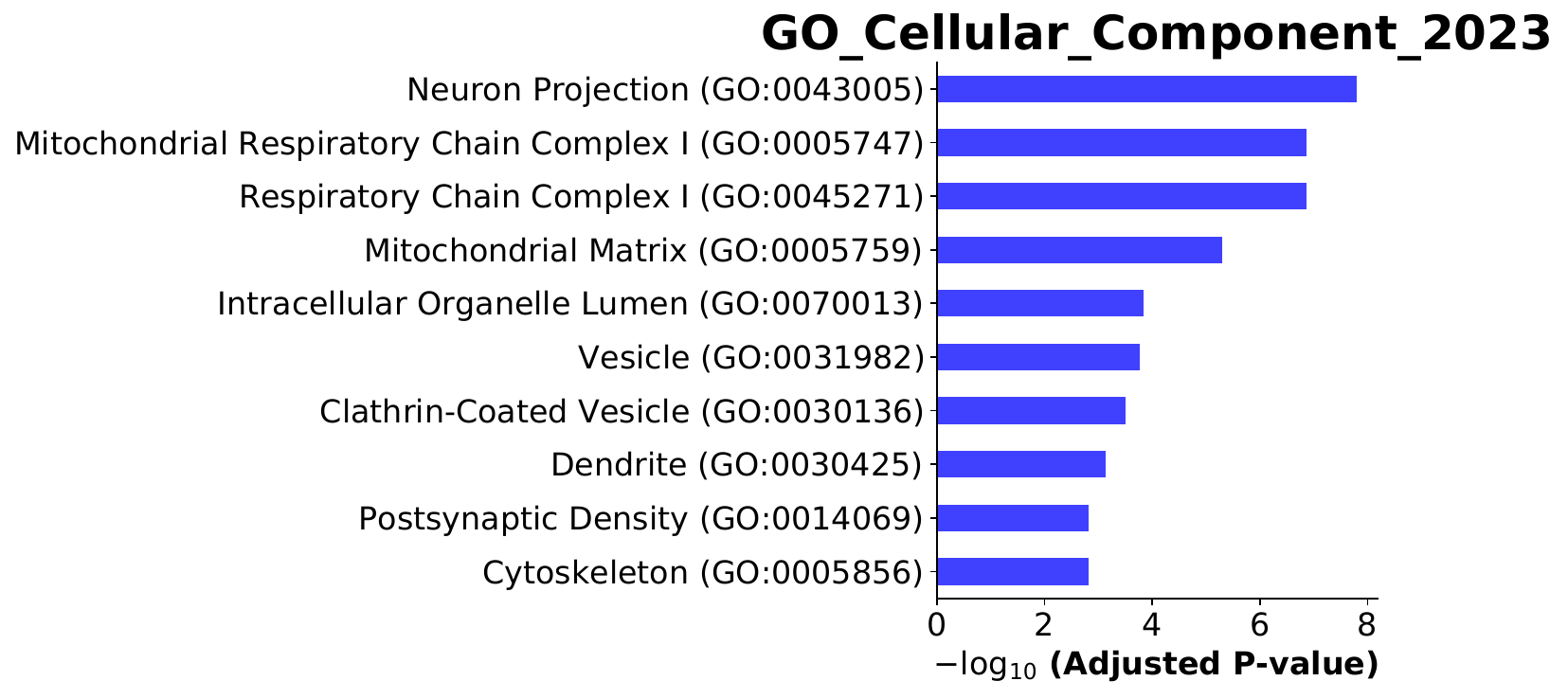}
\end{subfigure}
\hfill
\begin{subfigure}{0.5\textwidth}
\includegraphics[width=\linewidth, height=0.5\textwidth]{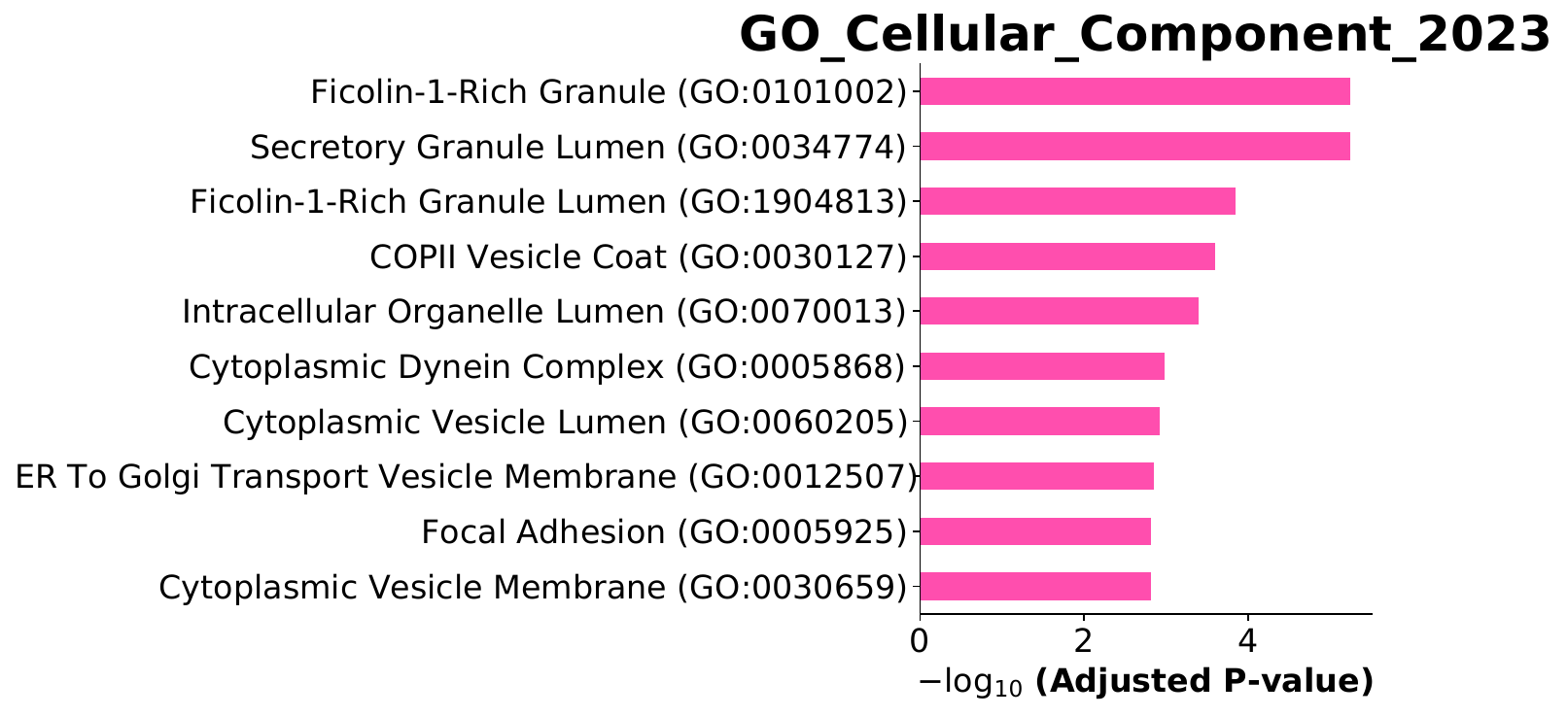}
\end{subfigure}

\caption{Gene Ontology enrichment analysis of proteins rhythmic in control subjects but not in AD (blue), and in AD subjects but not in control (pink) on temporal cortex.}
\label{fig:GSEA on TC}
\end{figure*}

 We conducted Gene Ontology (GO) enrichment analysis using GSEApy \cite{fang2023gseapy}, which serves as an interface between Python and Enrichr web services \cite{chen2013enrichr, kuleshov2016enrichr, xie2021gene}. This analysis was performed on rhythmic proteins in the control group that lose rhythmicity in  AD (control-specific rhythmic proteins), as well as rhythmic proteins in AD that are not rhythmic in the control group (AD-specific rhythmic proteins), for each brain region (Figures \ref{fig:GSEA on TC}, S12, and S14).

Observing the enriched biological processes in TC region (Figure \ref{fig:GSEA on TC}), both control-specific and AD-specific rhythmic proteins are enriched with respiration and energy metabolism.
For control-specific rhythmic proteins, the most significant process is aerobic respiration, whereas for AD-specific rhythmic proteins it is dicarboxylic acid catabolic process that is also related to energy metabolism. As intermediates, in the tricarboxylic acid cycle (TCA; also called citric acid cycle or Krebs cycle) such as succinate, fumarate, and malate, dicarboxylic acids are key substrates for cellular respiration. Their oxidation provides a critical source of electrons in the electron transport chain for oxidative phosphorylation and subsequent ATP synthesis through this integrated network of pathways.

The enrichment in AD-specific rhythmic proteins has distinctive processes such as response to endoplasmic reticulum stress, glutamate catabolic process and energy derivation by oxidation of organic compounds, which are not enriched in the control-specific rhythmic proteins. The response process is related to the activation of pathways managing stress affecting the endoplasmic reticulum (ER) in cells. Such stress is triggered by the accumulation of misfolded proteins in the ER lumen. The ER stress response involves signaling pathways that work to restore proper protein folding in the ER, manage the protein load, and degrade misfolded proteins. The unfolded protein response is one of the main pathways managing the cellular response to ER stress. It acts to restore ER homeostasis. If ER stress is prolonged, apoptotic cell death pathways may be activated. This enrichment aligns with literature on misfolded proteins in AD \cite{sweeney2017protein}.
 
The other processes enriched in AD-specific rhythmic proteins processes are linked to the energy metabolic alterations in AD. For example, glutamate catabolism connects tightly to multiple pathways related to fueling respiration, driving aerobic/anaerobic energy production, and contributing to nitrogen balance. Glutaminolysis pathway metabolically breaks down glutamate for substrates like lactate that is important for energy homeostasis, e.g., in cancer cells relying more on anaerobic glycolysis pathways. Our findings align with the results from the literature \cite{loehfelm2019timing}, which point out that circadian affects energy homeostasis and has significant effects on AD.

It is important to note that there is often an inherent biases in functional enrichment analysis. Our primary objective in using GSEA is not to establish definitive biological proof but rather to explore potential functional relevance of the rhythmic proteins. In this regard, the pathway analysis results should be viewed as a complementary  and exploratory perspective to our modeling process, offering additional insights into the biological context of rhythmic proteins.\color{black}
\subsubsection{Disease associated with control and AD specific rhythmic proteins}
We used DisGeNET \cite{pinero2015disgenet} library to find diseases associated with control-specific and AD-specific rhythmic proteins (see Figures S11(a), S13(a), and S15(a)). Schizophrenia, AD, and neurodegenerative disorders are shared among all rhythmic proteins in control, which lose rhythmicity in AD in all brain regions. This shows that disrupted proteins are highly associated with AD, as well as schizophrenia, which has a high correlation with AD \cite{chen2022genetic}. We also analyzed these proteins using additional annotation libraries (see Figures S11(b), S13(b), and S15(b)).

\begin{figure}[]
\centering
\begin{subfigure}{\textwidth}
\centering
\includegraphics[width=0.9\textwidth]{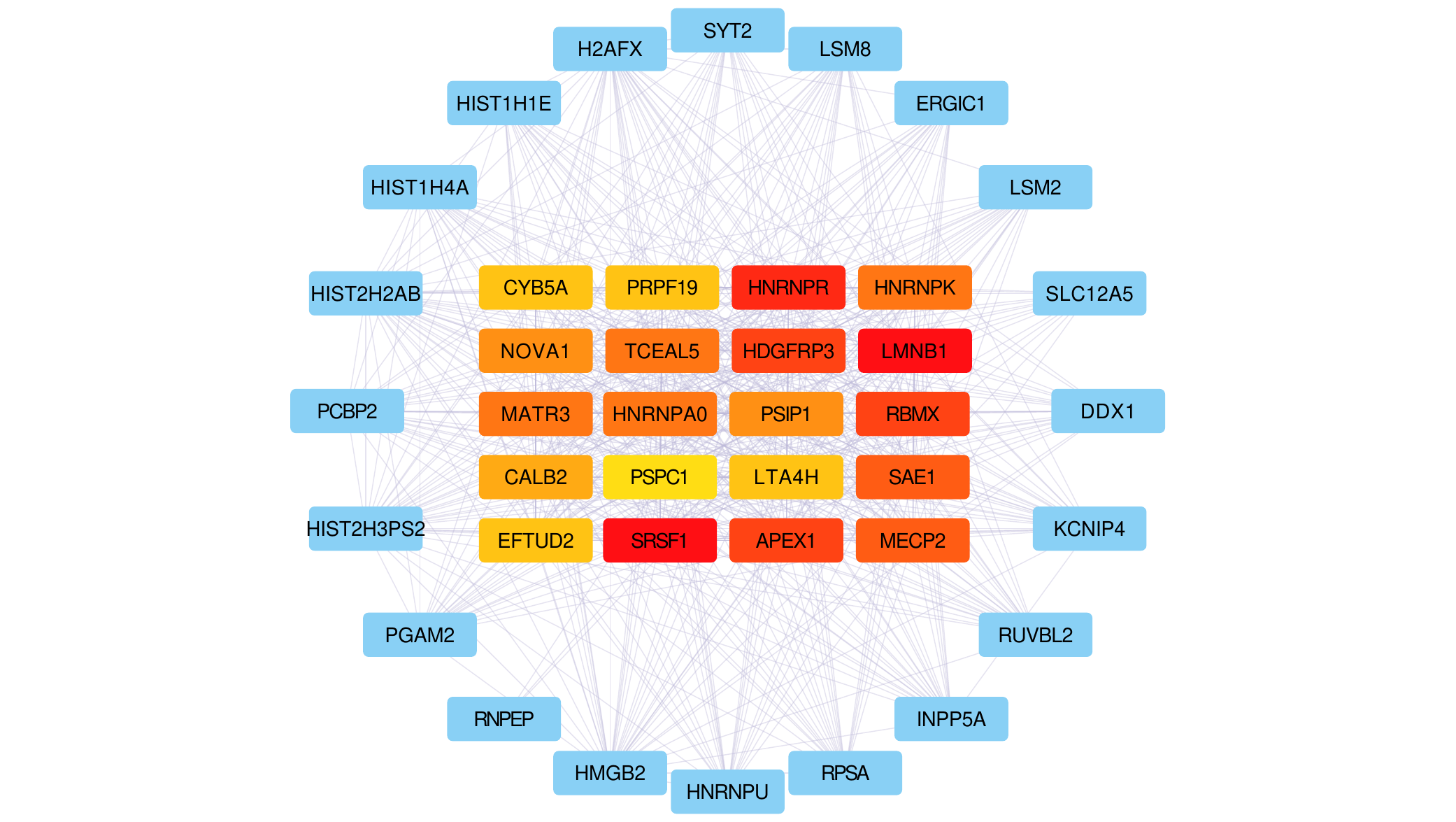}
\includegraphics[width=0.45\textwidth]{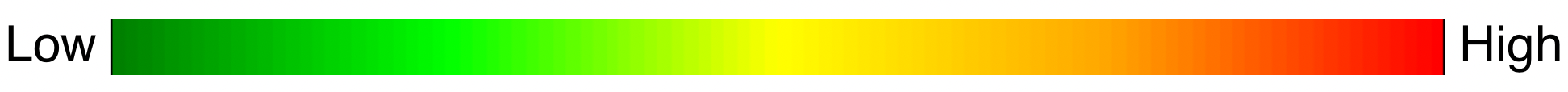}
\label{fig:net}
\caption{}
\end{subfigure}
    
\vspace{1em} 
    
\begin{subfigure}{\textwidth}
\centering
\small
\begin{tabular}{cc}
\rowcolor{lightblue}
\hline
\textbf{Hub Protein} & \textbf{Related Drug(s)} \\
\hline
LTA4H & Captopril [A], Ubenimex [I],
DG051 [I],\\& Tosedostat [I], Imidazole [E\&I] \\
\hline
SRSF1  & Copper [A\&I ] \\
\hline
HNRNPK & Artenimol [A\&E\&I], Phenethyl Isothiocyanate [I] \\
\hline
CYB5A & Chromium[A], Chromic nitrate [A],\\
& Chromium gluconate [A], Chromous sulfate [A],\\
 & Chromium nicotinate [A\&E] \\
\hline
CALB2 & Calcium Phosphate [A], Calcium phosphate dihydrate [A],\\ 
& Calcium citrate [A\&I], Calcium levulinate [A\&E ] \\
\hline
APEX1 & Lucanthone [I] \\
\hline
\end{tabular}
\label{tab:AD_drug}
\caption{}
\end{subfigure}
\caption{ Hub proteins in temporal cortex. (a) Network depiction of the first twenty hub proteins based on degrees and proteins directly interacting with these hub proteins. The colorbar indicates the connection degrees of the hub proteins. (b) Hub proteins and related drugs. The status of each drug is shown in the brackets next to it: A, I, and E indicate that the drug is approved, investigational, and experimental, respectively.}
\label{fig:hub proteins in TC}
\end{figure}
\subsubsection{Hub proteins and related drugs}
In addition, we present the top 20 hub proteins along with their directly connected first-stage nodes in the proteomic data of control-specific rhythmic proteins within the TC (Figure \ref{fig:hub proteins in TC}(a)). Similar analyses for the parietal association cortex, the DLPFC, and urine datasets are provided in Table S2. 
To achieve this, we initially constructed a protein coexpression network using the WGCNA R package \cite{langfelder2008wgcna}, focusing on proteins that display rhythmicity in the control group but lose rhythmicity in AD. We applied the cytoHubba Cytoscape plugin \cite{shannon2003cytoscape,chin2014cytohubba} and utilized the degree algorithm to identify and highlight the top-ranked proteins based on their connectivity with other proteins. Through the use of DrugBank \cite{wishart2006drugbank}, we discovered that six of these hub proteins represent potential drug targets. The corresponding drugs for each target are shown in Figure \ref{fig:hub proteins in TC}(b), categorized based on their development status as approved, investigational, or experimental.

\section{Discussion}

To study circadian rhythms at the protein level, sample times are needed. However, proteomic datasets, especially in humans, often lack explicit time labels. Other challenges include: the sample sizes are typically very small; ultradian proteins typically exist; and knowledge of rhythmic proteins are limited. To address these issues, PROTECT has been designed to predict sample phases in proteomic data in an unsupervised manner. This method is appropriate for different sample sizes and does not require a priori information. The effectiveness of PROTECT was validated through testing on time-labeled datasets, achieving remarkable results with over 80\% nAUC on all instances. Our exploration of un-labeled human datasets using PROTECT's time predictions revealed differences between control and AD subjects in 
three brain regions and urine samples. Additionally, the study provides potential drug targets and identifies some ultradian rhythmic proteins in TC. 

 PROTECT is generalizable to other circadian omics data types, such as transcriptomic and metabolomic datasets \cite{mauvoisin2014circadian,mure2018diurnal,zhang2014circadian,krishnaiah2017clock,vollmers2009time,hughes2009harmonics,talamanca2023sex,atger2015circadian,weger2021systematic}, as these data types share similar periodic structures. Unlike existing methods tailored for transcriptomics, PROTECT does not rely on time information or predefined rhythmic features, such as known rhythmic genes or metabolites, making it broadly applicable to these types of datasets. 

To validate this, we applied PROTECT to some publicly available circadian transcriptomic and metabolomic datasets. Specifically, we analyzed transcriptomic data from a mouse liver study \cite{mauvoisin2014circadian}, which includes both proteomic and transcriptomic measurements, as well as a baboon amygdala dataset \cite{mure2018diurnal} and a mouse kidney dataset \cite{zhang2014circadian}. Additionally, we evaluated metabolomic data from a mouse liver study \cite{krishnaiah2017clock}. PROTECT successfully predicted sample phases across these datasets, with detailed results provided in the Supplementary Material (Figure S19). These findings further support the robustness of our method and its potential for broader applications in circadian and diurnal research.\color{black}

\subsection{Rhythmic proteins in human dataset} On the TC and DLPFC, large portions of proteins are rhythmic in both control and AD subjects. However, in the parietal association cortex, only about 21\% of proteins show rhythmicity patterns. Moreover, in the urine dataset, 458, 434, and 387 out of 555 proteins show rhythmic patterns in control, MCI, and AD subjects, respectively (Figure S16). This indicates a significant portion of proteins exhibiting rhythmicity across all groups. However, it's important to note that the urine dataset had numerous missing values and after the removal of proteins with missing values in more than half of the samples, we lost many proteins. Therefore, we cannot determine whether they exhibit rhythmic patterns or not. Also, we observed different distributions of peak times in brain regions between control and AD subjects, which can also be seen in urine datasets (Figure S17). 

\subsection{Associated diseases} Our findings, focusing on proteins exhibiting rhythmicity in control subjects but losing this pattern in AD within the temporal cortex, identified mitochondrial disease as the primary associated condition. This observation is in line with previous studies \cite{bhatia2022mitochondrial, swerdlow2004mitochondrial} emphasizing the involvement of mitochondrial dysfunction in the progression of AD. Notably, our results also reveal schizophrenia as the second most associated disease across all brain regions. A study by \cite{chen2022genetic} supports our findings, demonstrating a robust genetic correlation between AD and schizophrenia.

\subsection{Hub proteins and drug targets} 
In our study of proteins exhibiting rhythmic behavior in control subjects but not in individuals with AD, we present the top 20 hub proteins in the TC. Notably, six of these proteins, namely LTA4H, SRSF1, HNRNPK, CYB5A, CALB2, and APEX1, emerge as potential drug targets. Recent studies have linked some of these proteins to be therapeutic targets for AD. According to Adams et al. (2023) \cite{adams2023leukotriene}, LTA4H has been recognized not only as a therapeutic target but also as a plasma biomarker for cognitive impairment associated with aging and AD. Additionally, SRSF1 and PTPB1 have been implicated in suppressing the formation of a CD33 splicing isoform associated with AD \cite{van2019srsf1}. Furthermore, there is evidence suggesting that CALB2, specifically in hippocampal interneurons, serves as an early target in a transgenic model exhibiting AD-like pathology \cite{baglietto2010calretinin}.

\subsection{Ultradian proteins} In the TC dataset, we utilized predicted sample times to identify potential ultradian rhythmic proteins. Our selection criteria included FDR $< 5e-4$, rAMP $\geq 0.2$, and $R^{2} \geq 0.6$ to select the proteins with a period of 12 hours. Through this analysis, we identified six proteins (ACOT7, CLTA, ELMO1, MDP1, NIT1, SH3GL2) exhibiting these characteristics, as illustrated in Figure S18.

\subsection{Potential limitations of our approach} Despite its excellent performance, PROTECT requires further exploration. In our study, we did not address the impact of the postmortem interval (PMI) on postmortem human brain datasets. PMI refers to the duration between the time of death and sample collection, which can influence proteomic values. While most samples used in this paper have a PMI of less than 10 hours, thus minimally impacting the study, future work should investigate the effect of PMI on predicted times. 

One limitation of PROTECT lies in its strategy for handling missing values within the proteomic dataset. The current practice of excluding proteins with missing values may lead to the exclusion of valuable data that could contribute to a more comprehensive understanding of our investigation. As advances in data collection lead to fewer missing values, we expect that more proteins can be incorporated into rhythmicity analysis.

Also, to determine which protein should exhibit its peak at time 0 and subsequently arrange the remaining proteins in relation to this target protein, we employed EHD1. This protein showed rhythmicity in all datasets for both control and AD subjects. We established the peak value of EHD1 at time 0 and organized the other proteins relative to EHD1. In the future, with more studies on proteins' peak times, we can use some known proteins as the target protein. 

Besides potential future enhancements to PROTECT itself, systematically investigating these limitations provides broader understanding of the data dependencies and robustness in circadian proteomic modeling. In turn, with further investigation into these practical data constraints alongside algorithmic advances, PROTECT may significantly influence standardized protocols for circadian omics analysis, thus enabling more consistent and accurate biological findings.

\section*{Data availability statement}
All data used in this paper are publicly available and cited in the paper.

\section*{CRediT authorship contribution statement}
\textbf{Aram Ansary Ogholbake:} Conceptualization, Formal Analysis, Investigation, Methodology, Software, Visualization, Writing – original draft
\textbf{Qiang Cheng:} Conceptualization, Funding acquisition, Investigation, Methodology, Supervision, Writing – review \& editing
\section*{Declaration of competing interest}
None Declared
\section*{Funding sources}
This work was supported by the by the NSF under Grants IIS 2327113 and ITE 2433190; and the NIH under Grants R21AG070909, P30AG072946, and R01HD101508-01.

\bibliographystyle{elsarticle-num}
\bibliography{reference}

\includepdf[pages=-]{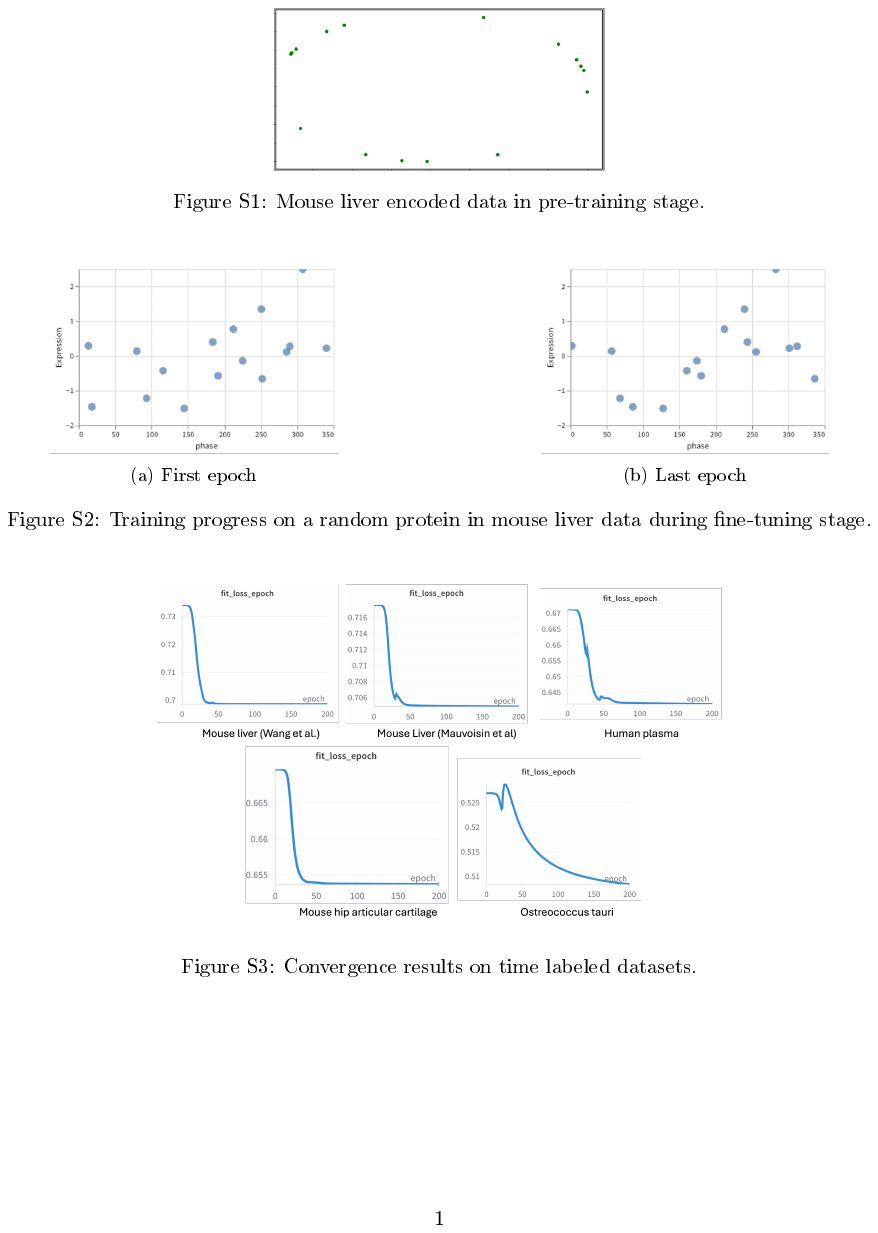}

\end{document}